\documentclass{article}

\usepackage[utf8]{inputenc}
\usepackage{amsmath, amsthm, amssymb, amsfonts, dsfont, mathrsfs, dirtytalk, tikz-cd, adjustbox, url, nicefrac, booktabs,array, graphicx,todonotes,longtable, tabu,bm,cite, todonotes,comment,multirow,tabularx,enumitem}
\usepackage[T1]{fontenc}    
\usepackage[english]{babel}
\usepackage[preprint=true]{jmlredit}

\newcommand{\R}{\mathbb R}

\newcommand{\E}{\mathbb E}

\newcommand{\A}{\mathbb A}
\renewcommand{\H}{\operatorname{H}}
\renewcommand{\S}{\mathbb{S}}
\newcommand{\T}{\mathbb{T}}

\begin{document}

\title{Reward Maximisation through Discrete Active Inference}

\author{\name Lancelot Da Costa \email l.da-costa@imperial.ac.uk \\
       \addr Department of Mathematics\\
       Imperial College London\\
       London, SW7 2AZ, UK
       \AND
       \name Noor Sajid \email noor.sajid.18@ucl.ac.uk \\
       \name Thomas Parr \email thomas.parr.12@ucl.ac.uk \\
       \name Karl Friston \email k.friston@ucl.ac.uk \\
       \addr Wellcome Centre for Human Neuroimaging\\
       University College London\\
       London, WC1N 3AR, UK
       \AND
       Ryan Smith \email rsmith@laureateinstitute.org \\
       \addr Laureate Institute for Brain Research\\
       Tulsa, OK 74136, United States
       }
       
\maketitle
\vspace{-25pt}
\begin{abstract}%
Active inference is a probabilistic framework for modelling the behaviour of biological and artificial agents, which derives from the principle of minimising free energy. In recent years, this framework has successfully been applied to a variety of situations where the goal was to maximise reward, offering comparable and sometimes superior performance to alternative approaches. In this paper, we clarify the connection between reward maximisation and active inference by demonstrating how and when active inference agents perform actions that are optimal for maximising reward. Precisely, we show the conditions under which active inference produces the optimal solution to the Bellman equation—a formulation that underlies several approaches to model-based reinforcement learning and control. On partially observed Markov decision processes, the standard active inference scheme can produce Bellman optimal actions for planning horizons of 1, but not beyond. In contrast, a recently developed recursive active inference scheme (sophisticated inference) can produce Bellman optimal actions on any finite temporal horizon. We append the analysis with a discussion of the broader relationship between active inference and reinforcement learning.
\end{abstract}

\tableofcontents

\section{Introduction}

\subsection{Active inference}

Active inference is a normative framework for modelling intelligent behaviour in biological and artificial agents. It simulates behaviour by numerically integrating equations of motion thought to describe the behaviour of biological systems, a description based on the free energy principle~\citep{fristonFreeEnergyPrinciple2022,barpGeometricMethodsSampling2022,fristonFreeenergyPrincipleUnified2010,ramsteadBayesianMechanicsPhysics2022}. Active inference comprises a collection of algorithms for modelling perception, learning, and decision-making in the context of both continuous and discrete state spaces \citep{fristonGraphicalBrainBelief2017,fristonSophisticatedInference2021,dacostaActiveInferenceDiscrete2020,buckleyFreeEnergyPrinciple2017,barpGeometricMethodsSampling2022,fristonActionBehaviorFreeenergy2010}. Briefly, building active inference agents entails: $1)$ equipping the agent with a (generative) model of the environment, $2)$ fitting the model to observations through approximate Bayesian inference by minimising variational free energy (i.e., optimising an evidence lower bound \citep{bealVariationalAlgorithmsApproximate2003,bishopPatternRecognitionMachine2006,jordanIntroductionVariationalMethods1998,bleiVariationalInferenceReview2017}) and $3)$ selecting actions that minimise expected free energy, a quantity that that can be decomposed into risk (i.e., the divergence between predicted and preferred paths) and ambiguity, leading to context-specific combinations of exploratory and exploitative behaviour \citep{schwartenbeckComputationalMechanismsCuriosity2019,millidgeApplicationsFreeEnergy2021}. This framework has been used to simulate and explain intelligent behaviour in neuroscience \citep{parrComputationalNeurologyMovement2021,adamsComputationalAnatomyPsychosis2013,parrComputationalNeurologyActive2019,sajidMixedGenerativeModel2022}, psychology and psychiatry~\citep{smithSlowerLearningRates2022,smithActiveInferenceApproach2021,smithGreaterDecisionUncertainty2021,smithLongtermStabilityComputational2021,smithImpreciseActionSelection2020a,smithBayesianComputationalModel2020,smithGutInferenceComputational2021,smithConfirmatoryEvidenceThat2020}, machine learning \citep{millidgeDeepActiveInference2020,tschantzReinforcementLearningActive2020,tschantzScalingActiveInference2019,fountasDeepActiveInference2020,catalLearningPerceptionPlanning2020a,mazzagliaContrastiveActiveInference2021} and robotics \citep{catalRobotNavigationHierarchical2021,lanillosRobotSelfOther2020,sancaktarEndtoEndPixelBasedDeep2020,pio-lopezActiveInferenceRobot2016,pezzatoNovelAdaptiveController2020,oliverEmpiricalStudyActive2021,schneiderActiveInferenceRobotic2022}.

\subsection{Reward maximisation through active inference?}

In contrast, the traditional approaches towards simulating and explaining intelligent behaviour---stochastic optimal control~\citep{bertsekasStochasticOptimalControl1996,bellmanDynamicProgramming1957} and reinforcement learning (RL; \citet{bartoReinforcementLearningIntroduction1992})---derive from the normative principle of executing actions to maximise reward scoring the utility afforded by each state of the world. This idea dates back to expected utility theory \citep{vonneumannTheoryGamesEconomic1944}, an economic model of rational choice behaviour, which also underwrites game theory \citep{vonneumannTheoryGamesEconomic1944} and decision theory \citep{dayanDecisionTheoryReinforcement2008,bergerStatisticalDecisionTheory1985}. Several empirical studies have shown that active inference can successfully perform tasks that involve collecting reward, often (but not always) showing comparative or superior performance to RL \citep{millidgeDeepActiveInference2020,paulActiveInferenceStochastic2021,cullenActiveInferenceOpenAI2018,vanderhimstDeepActiveInference2020,sajidActiveInferenceDemystified2021,markovicEmpiricalEvaluationActive2021a,mazzagliaContrastiveActiveInference2021,smithSlowerLearningRates2022,smithGreaterDecisionUncertainty2021,smithLongtermStabilityComputational2021,smithImpreciseActionSelection2020a}, and marked improvements when interacting with volatile environments \citep{sajidActiveInferenceDemystified2021,markovicEmpiricalEvaluationActive2021a}. Given the prevalence and historical pedigree of reward maximisation, we ask:

\emph{How and when do active inference agents execute actions that are optimal with respect to reward maximisation?}

\subsection{Organisation of paper}

In this paper, we explain (and prove) how and when active inference agents exhibit (Bellman) optimal reward maximising behaviour.

For this, we start by restricting ourselves to the simplest problem: maximising reward on a finite horizon Markov decision process (MDP) with known transition probabilities---a sequential decision-making task with complete information. In this setting, we review the backward induction algorithm from dynamic programming, which forms the workhorse of many optimal control and model-based RL algorithms. This algorithm furnishes a Bellman optimal state-action mapping, which means that it provides provably optimal decisions from the point of view of reward maximisation (Section \ref{sec: Dynamic programming on finite horizon MDPs}). 

We then introduce active inference on finite horizon MDPs (Section \ref{sec: Active inference on finite horizon MDPs})---a scheme consisting of perception as inference followed by planning as inference, which selects actions so that future states best align with preferred states.

In Section \ref{sec: reward maximisation}, we show how and when active inference maximises reward in MDPs. Specifically, when the preferred distribution is a (uniform mixture of) Dirac distribution(s) over reward maximising trajectories, selecting \textit{action sequences} according to active inference maximises reward (Section \ref{sec: EFE MDP}). Yet, active inference agents, in their standard implementation, can select \textit{actions} that maximise reward \textit{only} when planning one step ahead (Section \ref{sec: AI MDP}). It takes a recursive, \textit{sophisticated} form of active inference to select actions that maximise reward---in the sense of a Bellman optimal state-action mapping---on any finite time-horizon (Section \ref{sec: SI MDP}).

In Section \ref{sec: generalisation to POMDPs}, we introduce active inference on partially observable Markov decision processes with known transition probabilities---a sequential decision-making task where states need to be inferred from observations---and explain how the results from the MDP setting generalise to this setting.

Our findings are summarised in Section \ref{sec: conclusion}.

All of our analyses assume that the agent knows the environmental dynamics (i.e., transition probabilities) and reward function. In Appendix \ref{app: AIF and RL}, we discuss how active inference agents can learn their world model and rewarding states when these are initially unknown---and the broader relationship between active inference and RL.

\section{Reward maximisation on finite horizon MDPs}
\label{sec: Dynamic programming on finite horizon MDPs}
In this section, we consider the problem of reward maximisation in Markov decision processes (MDPs) with known transition probabilities.

\subsection{Basic definitions}
MDPs are a class of models specifying environmental dynamics widely used in dynamic programming, model-based RL, and more broadly in engineering and artificial intelligence~\citep{bartoReinforcementLearningIntroduction1992,stoneArtificialIntelligenceEngines2019}. They are used to simulate sequential decision-making tasks with the objective of maximising a reward or utility function. An MDP specifies environmental dynamics unfolding in discrete space and time under the actions pursued by an agent.

\begin{figure}[t!]
    \centering
    \includegraphics[width= 250pt]{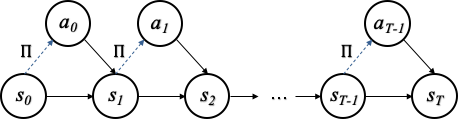}
    \caption{\textbf{Finite horizon Markov decision process.} This is a Markov decision process pictured as a Bayesian network~\citep{pearlGraphicalModelsProbabilistic1998,jordanIntroductionVariationalMethods1998}. A finite horizon MDP comprises a finite sequence of states, indexed in time. The transition from one state to the next state depends on action. As such, for any given action sequence, the dynamics of the MDP form a Markov chain on state-space. In this fully observed setting, actions can be selected under a state-action policy, $\Pi$, indicated with a dashed line: this is a probabilistic mapping from state-space and time to actions.}
    \label{fig: MDP}
\end{figure}

\begin{definition}[Finite horizon MDP]
\label{def: MDP}
A finite horizon MDP comprises the following collection of data:
\begin{itemize}
    \item $\mathbb S$ a finite set of states.
    \item $\mathbb T= \{0,...,T\}$ a finite set which stands for discrete time. $T$ is the temporal horizon (a.k.a. planning horizon).
    \item $\mathbb A$ is a finite set of actions.
    \item \(P(s_t=s'\mid s_{t -1}=s, a_{t-1}=a)\) is the probability that action \(a \in \mathbb A\) in state \(s \in \mathbb S\) at time \(t-1\) will lead to state \(s^{\prime} \in \mathbb S\) at time \(t\). $s_{t}$ are random variables over $\mathbb S$ that correspond to the state being occupied at time $t=0,...,T$.
    \item $P(s_0=s)$ specifies the probability of being at state $s \in \mathbb S$ at the start of the trial.
    \item $R(s)$ is the finite reward received by the agent when at state $s \in \S$. 
\end{itemize}
The dynamics afforded by a finite horizon MDP (see Figure \ref{fig: MDP}) can be written globally as a probability distribution over state trajectories $s_{0:T}:=(s_0,\ldots, s_T)$, given a sequence of actions $a_{0:T-1}:=(a_0,\ldots, a_{T-1})$, which factorises as follows:
    \begin{equation*}
    \begin{split}
    P(s_{0:T}\mid a_{0:T-1}) &=P(s_0) \prod_{\tau =1}^T P(s_\tau\mid s_{\tau -1}, a_{\tau-1}).
    \end{split}
\end{equation*}

\end{definition}

\begin{remark}[On the definition of reward]
More generally, the reward function can be taken to be dependent on the previous action and previous state: \(R_{a}\left(s^{\prime}\mid s\right)\) is the reward received after transitioning from state \(s\) to state \(s^{\prime},\) due to action \(a\)~\citep{bartoReinforcementLearningIntroduction1992,stoneArtificialIntelligenceEngines2019}. However, given an MDP with such a reward function, we can recover our simplified setting by defining a new MDP where the new states comprise the previous action, previous state, and current state in the original MDP. By inspection, the resulting reward function on the new MDP depends only on the current state (i.e., $R(s)$).
\end{remark}

\begin{remark}[Admissible actions]
In general, it is possible that only \emph{some} actions can be taken at each state. In this case, one defines \(\mathbb A_{s}\) to be the finite set of (allowable) actions from state \(s \in \S\). All forthcoming results concerning MDPs can be extended to this setting.
\end{remark}

To formalise what it means to choose actions in each state, we introduce the notion of a state-action policy.

\begin{definition}[State-action policy]
    A state-action policy $\Pi$ is a probability distribution over actions, that depends on the state that the agent occupies, and time. Explicitly,
    \begin{equation*}
    \begin{split}
        \Pi: \mathbb A \times \S \times \T &\to [0,1] \\
        (a,s ,t) &\mapsto \Pi(a \mid s,t) \\
       \forall (s,t) \in \S \times \T&: \sum_{a\in A} \Pi(a\mid s,t) =1.
    \end{split}
    \end{equation*}
    When $s_t =s$, we will write $\Pi(a\mid s_t) := \Pi(a\mid s,t)$. Note that the action at the temporal horizon $T$ is redundant, as no further reward can be reaped from the environment. Therefore, one often specifies state-action policies only up to time $T-1$, as $\Pi: \mathbb A \times \S \times \{0,\ldots, T-1\} \to [0,1]$. The state-action policy---as defined here---can be regarded as a generalisation of a deterministic state-action policy that assigns the probability of $1$ to an available action and $0$ otherwise. 
\end{definition}

\begin{remark}[Conflicting terminologies: policy in active inference]
In active inference, a \emph{policy} is defined as a sequence of actions indexed in time\footnote{These are analogous to temporally extended actions or \emph{options} introduced under the options framework in RL~\citep{stolle2002learning}.}. To avoid terminological confusion, we use \emph{action sequences} to denote policies under active inference.
\end{remark}

At time $t$, the goal is to select an action that maximises future cumulative reward:
\begin{equation*}
    R(s_{t+1:T}) := \sum_{\tau=t+1}^T R(s_\tau).
\end{equation*}

Specifically, this entails following a state-action policy $\Pi$ that maximises the \textit{state-value function}:
\begin{equation*}
\begin{split}
       v_\Pi(s,t) &:= \E_{\Pi}[ R(s_{t+1:T})\mid s_t=s]
\end{split}
\end{equation*}

for any $(s,t) \in \S \times \T$. The state-value function scores the expected cumulative reward if the agent pursues state-action policy $\Pi$ from the state $s_t=s$. When the state $s_t =s$ is clear from context, we will often write $v_\Pi(s_t) := v_\Pi(s,t)$. Loosely speaking, we will call the expected reward the \emph{return}.

\begin{remark}[Notation $\E_\Pi$]
\label{rem: notation expectation}
Whilst standard in RL~\citep{bartoReinforcementLearningIntroduction1992,stoneArtificialIntelligenceEngines2019}, the notation $\E_{\Pi}[ R(s_{t+1:T})\mid s_t=s]$
can be confusing. It denotes the expected reward, under the transition probabilities of the MDP and a state-action policy $\Pi$
\begin{equation*}
  \E_{P(s_{t+1:T}\mid a_{t:T-1},s_t=s)\Pi(a_{t:T-1}\mid s_{t+1:T-1},s_t=s)}[ R(s_{t+1:T})].
\end{equation*}
It is important to keep this correspondence in mind, as we will use both notations depending on context.
\end{remark}

\begin{remark}[Temporal discounting]
In infinite horizon MDPs (i.e., when $T$ is infinite), RL often seeks to maximise the discounted sum of rewards
\begin{equation*}
\begin{split}
       v_\Pi(s,t) &:= \E_{\Pi}\left[ \sum_{\tau=t}^\infty \gamma^{\tau-t} R(s_{\tau+1})\mid s_t=s\right],
\end{split}
\end{equation*}
for a given temporal discounting term $\gamma \in (0,1)$~\citep{bertsekasStochasticOptimalControl1996,bartoReinforcementLearningIntroduction1992,kaelblingPlanningActingPartially1998}. In fact, temporal discounting is added to ensure that the infinite sum of future rewards converges to a finite value~\citep{kaelblingPlanningActingPartially1998}. In finite horizon MDPs temporal discounting is not necessary so we set $\gamma=1$ (c.f., ~\citep{schmidhuberFormalTheoryCreativity2010,schmidhuberDevelopmentalRoboticsOptimal2006}).
\end{remark}

To find the best state-action policies, we would like to rank them in terms of their return. We introduce a partial ordering  such that a state-action policy is \emph{better} than another if it yields a higher return in any situation:
\begin{equation*}
    \Pi \geq \Pi^\prime \iff \forall (s,t)\in \S\times \mathbb T: v_\Pi(s,t) \geq v_{\Pi^\prime}(s,t).
\end{equation*}
Similarly, a state-action policy $\Pi$ is \emph{strictly better} than another $\Pi'$ if it yields strictly higher returns:
\begin{equation*}
    \Pi > \Pi^\prime \iff \Pi \geq  \Pi^\prime \text{ and } \exists (s,t)\in \S\times \mathbb T: v_\Pi(s,t) > v_{\Pi^\prime}(s,t).
\end{equation*}

\subsection{Bellman optimal state-action policies}

A state-action policy is Bellman optimal if it is better than all alternatives. 

\begin{definition}[Bellman optimality]
A state-action policy $\Pi^*$ is \emph{Bellman optimal} if and only if it is better than all other state-action policies:
\begin{equation*}
    \Pi ^* \geq \Pi , \forall \Pi.
\end{equation*}
In other words, it maximises the state-value function $v_\Pi (s,t)$ for any state $s$ at time $t$.
\end{definition}

It is important to verify that this concept is not vacuous.

\begin{proposition}[Existence of Bellman optimal state-action policies]
\label{prop: Existence of the Bellman optimal state-action policy}
Given a finite horizon MDP as specified in Definition \ref{def: MDP}, there exists a Bellman optimal state-action policy $\Pi^*$. 
\end{proposition}

A proof can be found in Appendix \ref{app: proof 1}. Note that uniqueness of the Bellman optimal state-action policy is not implied by Proposition \ref{prop: Existence of the Bellman optimal state-action policy}; indeed, multiple Bellman optimal state-action policies may exist~\citep{putermanMarkovDecisionProcesses2014,bertsekasStochasticOptimalControl1996}.

Now that we know that Bellman optimal state-action policies exist, we can characterise them as a return-maximising action followed by a Bellman optimal state-action policy. 

\begin{proposition}[Characterisation of Bellman optimal state-action policies]
\label{prop: charac Bellman optimal state-action policy}
For a state-action policy $\Pi$, the following are equivalent:
\begin{enumerate}
    \item $\Pi$ is Bellman optimal.
    \item $\Pi$ is both
        \begin{enumerate}
            \item Bellman optimal when restricted to $\{1,\ldots, T\}$. In other words, $\forall$ state-action policy $\Pi'$ and $(s,t) \in \S \times \{1,\ldots T\}$
            \begin{equation*}
                v_\Pi (s,t ) \geq v_{\Pi'} (s,t ).
            \end{equation*}
            \item At time $0$, $\Pi$ selects actions that maximise return:
            \begin{equation}
            \label{eq: characterisation of Bellman optimal state-action policy}
            \Pi(a\mid s,0) > 0 \iff a \in \arg\max_{a\in \A} \E_{\Pi}[R(s_{1:T})\mid s_0=s,a_0=a] , \quad \forall s \in \S.
            \end{equation}
        \end{enumerate}
\end{enumerate}
\end{proposition}

A proof can be found in Appendix \ref{app: proof 2}. Note that this characterisation offers a recursive way to construct Bellman optimal state-action policies by successively selecting the best action, as specified by \eqref{eq: characterisation of Bellman optimal state-action policy}, starting from $T$ and inducting backwards~\citep{putermanMarkovDecisionProcesses2014}.

\subsection{Backward induction}

Proposition \ref{prop: charac Bellman optimal state-action policy} suggests a straightforward recursive algorithm to construct Bellman optimal state-action policies known as \emph{backward induction}~\citep{putermanMarkovDecisionProcesses2014}. Backward induction has a long history. It was developed by the German mathematician Zermelo in 1913 to prove that chess has Bellman optimal strategies~\citep{zermeloUberAnwendungMengenlehre1913}. In stochastic control, backward induction is one of the main methods for solving the Bellman equation~\citep{mirandaAppliedComputationalEconomics2002,addaDynamicEconomicsQuantitative2003,sargentOptimalControl2000}. In game theory, the same method is used to compute sub-game perfect equilibria in sequential games~\citep{fudenbergGameTheory1991}.

Backward induction entails planning backwards in time, from a goal state at the end of a problem, by recursively determining the sequence of actions that enables reaching the goal. It proceeds by first considering the last time at which a decision might be made and choosing what to do in any situation at that time in order to get to the goal state. Using this information, one can then determine what to do at the second-to-last decision time. This process continues backwards until one has determined the best action for every possible situation or state at every point in time.

\begin{proposition}[Backward induction: construction of Bellman optimal state-action policies]
\label{prop: backward induction construction of Bellman optimal state-action policies}
Backward induction
\begin{equation}
\label{eq: construction of Bellman optimal state-action policies}
\begin{split}
 \Pi(a\mid s,T-1) >0 \iff a &\in \arg\max_{a \in \A} \E[R(s_{T})\mid s_{T-1}=s,a_{T-1}=a], \quad \forall s\in \mathbb S \\
   \Pi(a\mid s,T-2) >0 \iff a &\in \arg\max_{a \in \A} \E_{\Pi}[R(s_{T-1:T})\mid s_{T-2}=s,a_{T-2}=a], \quad \forall s\in \mathbb S \\
    &\vdots\\
  \Pi(a\mid s,0) >0 \iff a &\in \arg\max_{a \in \A} \E_{\Pi}[R(s_{1:T})\mid s_{0}=s,a_{0}=a] , \quad \forall s\in \mathbb S\\
  \end{split}
\end{equation}
defines a Bellman optimal state-action policy $\Pi$. Furthermore, this characterisation is complete: all Bellman optimal state-action policies satisfy the backward induction relation \eqref{eq: construction of Bellman optimal state-action policies}.
\end{proposition}

A proof can be found in Appendix \ref{app: proof 3}.

\begin{example}[Intuition for backward induction]
To give a concrete example of this kind of planning, backward induction \eqref{eq: construction of Bellman optimal state-action policies} would consider the actions below in the following order:
\begin{enumerate}
    \item Desired goal: I would like to go to the grocery store,
    \item Intermediate action: I need to drive to the store,
    \item Current best action: I should put my shoes on.
\end{enumerate}
\end{example}

Proposition \ref{prop: backward induction construction of Bellman optimal state-action policies} tells us that to be optimal with respect to reward maximisation, one must plan like backward induction. This will be central to our analysis of reward maximisation in active inference.

\section{Active inference on finite horizon MDPs}
\label{sec: Active inference on finite horizon MDPs}

We now turn to introducing active inference agents on finite horizon MDPs with known transition probabilities.
We assume that the agent's generative model of its environment is given by the previously defined finite horizon MDP (see Definition \ref{def: MDP}). We do not consider the case where the transitions have to be learned but comment on it in the Appendix \ref{app: reward learning} (see also~\citep{dacostaActiveInferenceDiscrete2020,fristonActiveInferenceLearning2016}).

In what follows, we fix a time $t\geq 0$ and suppose that the agent has been in states $s_0,\ldots,s_t$. To ease notation, we let $\vec s := s_{t+1:T}, \vec a := a_{t:T}$ be the future states and future actions. We define $Q$ to be the \textit{predictive distribution}, which encodes the predicted future states and actions given that the agent is in state $s_t$
\begin{equation*}
     Q(\vec s,\vec a\mid s_t):=\prod_{\tau=t}^{T-1} Q(s_{\tau +1}\mid a_\tau, s_{\tau})Q(a_\tau\mid s_\tau).
\end{equation*}

\subsection{Perception as inference}

In active inference, perception entails inferences about future, past, and current states given observations and a sequence of actions. When states are partially observed, this is done through variational Bayesian inference by minimising a free energy functional (a.k.a. an evidence bound~\citep{bishopPatternRecognitionMachine2006,bealVariationalAlgorithmsApproximate2003,wainwrightGraphicalModelsExponential2007,bleiVariationalInferenceReview2017}).

In the MDP setting, past and current states are known, so it is only necessary to infer future states given the current state and action sequence $P(\vec s\mid \vec a,s_t)$. These posterior distributions $P(\vec s\mid \vec a,s_t)$ can be computed exactly in virtue of the fact that the transition probabilities of the MDP are known; hence variational inference becomes exact Bayesian inference.

\begin{equation}
\label{eq: transition probabilities are known}
\begin{split}
    Q(\vec s\mid \vec a,s_t)&:= P(\vec s\mid \vec a,s_t)= \prod_{\tau =t}^{T-1} P(s_{\tau+1}\mid s_{\tau }, a_{\tau}).
\end{split}
\end{equation}

\subsection{Planning as inference}
Now that the agent has inferred future states given alternative action sequences, we must assess these alternative plans by examining the resulting state trajectories. The objective that active inference agents optimise---in order to select the best possible actions---is the \textit{expected free energy}~\citep{barpGeometricMethodsSampling2022,dacostaActiveInferenceDiscrete2020,fristonSophisticatedInference2021}. Under active inference, agents minimise expected free energy in order to maintain themselves distributed according to a target distribution $C$ over the state-space $\S$ encoding the agent's preferences.

\begin{definition}[Expected free energy on MDPs]
On MDPs, the expected free energy of an action sequence $\vec a$ starting from $s_t$ is defined as~\citep{barpGeometricMethodsSampling2022}:
\begin{equation}
\label{eq: EFE def}
   G(\vec a\mid s_t)= \operatorname{D_{KL}}[Q(\vec s\mid \vec a,s_t)\mid C(\vec s)] 
\end{equation}
Therefore, minimising expected free energy corresponds to making the distribution over predicted states close to the distribution $C$ that encodes prior preferences. Note that the expected free energy in \emph{partially observed MDPs} comprises an additional ambiguity term (see Section \ref{sec: generalisation to POMDPs}), which is dropped here as there is no ambiguity about observed states.
\end{definition}

Since the expected free energy assesses the goodness of inferred future states under a course of action, we can refer to planning as inference~\citep{attiasPlanningProbabilisticInference2003,botvinickPlanningInference2012}.
The expected free energy may be rewritten as
\begin{equation}
   G(\vec a\mid s_t)= \underbrace{\E_{Q(\vec s\mid \vec a,s_t)}[-\log C(\vec s)]}_{\text{Expected surprise}}- \underbrace{\H[Q(\vec s\mid \vec a,s_t)]}_{\text{Entropy of future states}} 
\end{equation}
Hence, minimising expected free energy minimises the expected surprise of states\footnote{The surprise (a.k.a. self information or surprisal) of states $-\log C(\vec s)$ is information theoretic nomenclature~\citep{stoneInformationTheoryTutorial2015} that scores the extent to which an observation is unusual under $C$. It does not imply that the agent experiences surprise in a subjective or declarative sense.} according to $C$ and maximises the entropy of Bayesian beliefs over future states (a maximum entropy principle~\citep{jaynesInformationTheoryStatistical1957}, which is sometimes cast as keeping options open~\citep{klyubinKeepYourOptions2008}).

\begin{remark}[Numerical tractability]
The expected free energy is straightforward to compute using linear algebra. Given an action sequence $\vec a$, $C(\vec s)$ and $Q(\vec s\mid \vec a,s_t)$ are categorical distributions over $\mathbb S^{T-t}$. Let their parameters be $\textbf c, \textbf s_{\vec a} \in [0,1]^{\mid \mathbb S\mid (T-1)}$, where $\mid \cdot\mid $ denotes the cardinality of a set. Then the expected free energy reads
\begin{equation}
\label{eq: efe computational}
\begin{split}
    G(\vec a\mid s_t)&= \textbf s_{\vec a}^{\operatorname{T}} (\log \textbf s_{\vec a} -\log \textbf c).
\end{split}
\end{equation}
Notwithstanding, \eqref{eq: efe computational} is expensive to evaluate repeatedly when all possible action sequences are considered. In practice, one can adopt a temporal mean field approximation over future states~\citep{millidgeWhenceExpectedFree2020}:

\begin{equation*}
    Q(\vec s\mid \vec a,s_t) \approx \prod_{\tau=t+1}^T Q(s_\tau\mid \vec a,s_t),
\end{equation*}
which yields the simplified expression
\begin{equation}
\label{eq: EFE temporal mean field}
    G(\vec a\mid s_t)\approx \sum_{\tau =t+1}^T \operatorname{D_{KL}}[Q(s_\tau\mid \vec a,s_t)\mid C(s_\tau)].
\end{equation}

Expression \eqref{eq: EFE temporal mean field} is much easier to handle: for each action sequence $\vec a$, 1) one evaluates the summands sequentially $\tau= t+1, \ldots, T$, and 2) if and when the sum up to $\tau$ becomes significantly higher than the lowest expected free energy encountered during planning, $G(\vec a\mid s_t)$ is set to an arbitrarily high value. Setting $G(\vec a\mid s_t)$ to a high value is equivalent to pruning away unlikely trajectories. This bears some similarity to decision tree pruning procedures used in RL~\citep{huysBonsaiTreesYour2012}. It finesses exploration of the decision-tree in full depth and provides an Occam's window for selecting action sequences.

Complementary approaches can help make planning tractable. For example, hierarchical generative models factorise decisions into multiple levels. By abstracting information at a higher-level, lower-levels entertain fewer actions~\citep{fristonDeepTemporalModels2018}---which reduces the depth of the decision tree by orders of magnitude. Another approach is to use algorithms that search the decision-tree selectively, such as Monte-Carlo tree search~\citep{silverMasteringGameGo2016,championBranchingTimeActive2021,fountasDeepActiveInference2020,maistoActiveTreeSearch2021,championBranchingTimeActive2021a}, and amortising planning using artificial neural networks (i.e., learning to plan)~\citep{catalLearningPerceptionPlanning2020a,fountasDeepActiveInference2020,sajidExplorationPreferenceSatisfaction2021}.
\end{remark}

\section{Reward maximisation on MDPs through active inference}\label{sec: reward maximisation}

Here, we show how active inference solves the reward maximisation problem.

\subsection{Reward maximisation as reaching preferences}
\label{sec: EFE MDP}

From the definition of expected free energy \eqref{eq: EFE def}, active inference on MDPs can be thought of as reaching and remaining at a target distribution $C$ over state-space. 

The idea that underwrites this section is that when the stationary distribution has all of its mass on reward maximising states, the agent will maximise reward. To illustrate this, we define a preference distribution $C_\beta, \beta >0$ over state-space $\S$, such that preferred states are rewarding states\footnote{Note the connection with statistical mechanics: $\beta$ is an inverse temperature parameter, $-R$ is a potential function and $C_\beta$ is the corresponding Gibbs distribution~\citep{pavliotisStochasticProcessesApplications2014,rahmeTheoreticalConnectionStatistical2019}.} 

\begin{equation*}
\begin{split}
    C_\beta(\sigma) &:= \frac{\exp \beta R(\sigma)}{\sum_{\varsigma \in \S}\exp \beta R(\varsigma)} \propto \exp (\beta R(\sigma)),  \quad \forall \sigma \in \S\\
  \iff  -\log C_\beta(\sigma) &= -\beta R(\sigma) - c(\beta), \quad \forall \sigma \in \S,  \text{ for some } c(\beta) \in \R \text{ constant w.r.t } \sigma.
\end{split}
\end{equation*}

The (inverse temperature) parameter $\beta>0$ scores how motivated the agent is to occupy reward maximising states. Note that states $s \in \S$ that maximise the reward $R(s)$ maximise $C_\beta(s)$ and minimise $-\log C_\beta(s)$ for any $\beta>0$.

Using the additive property of the reward function, we can extend $C_\beta$ to a probability distribution over trajectories $\vec \sigma := (\sigma_1,\ldots , \sigma_T)\in  \S^T$. Specifically, $C_\beta $ scores to what extent a trajectory is preferred over another trajectory:

\begin{equation}
\label{eq: defining preferences as reward maximising}
\begin{split}
    C_\beta(\vec \sigma) &:= \frac{\exp \beta R(\vec \sigma)}{\sum_{\vec \varsigma \in \S^T}\exp \beta R(\vec \varsigma)} = \prod_{\tau=1}^T \frac{\exp\beta R( \sigma_\tau)}{\sum_{\varsigma \in \S}\exp \beta R(\varsigma)}= \prod_{\tau=1}^T C_\beta(\sigma_\tau ),  \quad \forall \vec \sigma \in \S^T\\
  \iff  &-\log C_\beta(\vec \sigma) = -\beta R(\vec \sigma) - c'(\beta) = -\sum_{\tau=1}^T \beta R(\sigma_\tau)-c'(\beta), \quad \forall \vec \sigma \in \S^T,
\end{split}
\end{equation}
where $c'(\beta):= c(\beta)T \in \R$ is constant w.r.t $\vec \sigma$.

When the preferences are defined in this way, the zero-temperature limit $\beta \to +\infty$ is the case where the preferences $C_\beta$ are non-zero only for states or trajectories that maximise reward. In this case, $\lim_{\beta \to +\infty} C_\beta$ is a uniform mixture of Dirac distributions over reward maximising trajectories:
\begin{equation}
\label{eq: zero temp limit of preferences}
\begin{split}
    \lim_{\beta \to +\infty} C_\beta&\propto \sum_{\vec s \in I^{T-t}} \operatorname{Dirac}_{\vec s}\\
        I&:= \arg\max_{s\in \S} R(s).
\end{split}
\end{equation}
This is because, for a reward maximising state $\sigma$, $\exp(\beta R(\sigma))$ will converge to $+\infty$ more quickly than $\exp(\beta R( \sigma'))$ for a non-reward maximising state $\sigma'$. Since $C_\beta$ is constrained to be normalised to $1$ (as it is a probability distribution), $C_\beta(\sigma') \xrightarrow{\beta \to +\infty} 0$. Hence, in the limit $\beta \to +\infty$, $C_\beta$ is non-zero (and uniform) only on reward maximising states.

We now show how reaching preferred states can be formulated as reward maximisation:

\begin{lemma}
\label{lemma: reward maxim EFE minim zero temp}
The sequence of actions that minimises expected free energy also maximises expected reward in the zero temperature limit $\beta \to +\infty$ \eqref{eq: zero temp limit of preferences}:
\begin{equation*}
    \lim_{\beta \to +\infty} \arg\min_{\vec a } G(\vec a\mid s_t) \subseteq \arg\max_{\vec a} \E_{Q(\vec s\mid \vec a , s_t)}[R(\vec s)]
\end{equation*}
Furthermore, of those action sequences that maximise expected reward, the expected free energy minimisers will be those that maximise the entropy of future states $\H[Q(\vec s\mid \vec a, s_t)]$.
\end{lemma}

\begin{figure}[t!]
    \centering
    \includegraphics[width=0.6\textwidth]{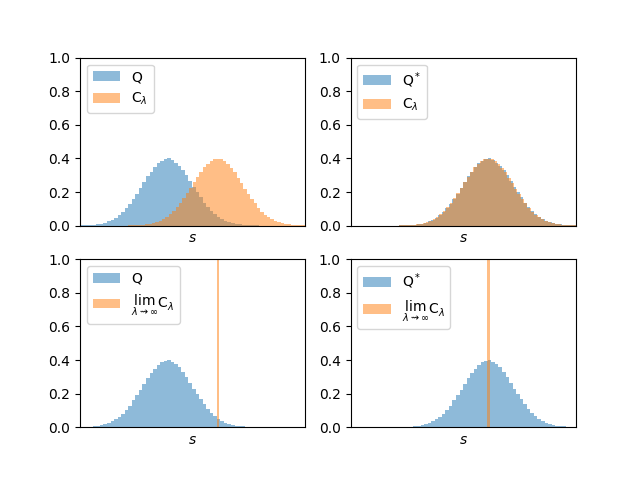}
    \caption{\textbf{Reaching preferences and the zero temperature limit.} We illustrate how active inference selects actions such that $Q(\vec s\mid \vec a,s_t)$ most closely matches the preference distribution $C_\beta$ (top-right). In this example, the discrete state-space is a discretisation of a continuous interval in $\R$, and the preferences and predictive distributions over states have a Gaussian shape. The predictive distribution $Q$ is assumed to have a fixed variance with respect to action sequences, such that the only parameter that can be optimised by action selection is its mean. Crucially, in the zero temperature limit \eqref{eq: zero temp limit of preferences}, $\lim_{\beta \to +\infty} C_\beta$ becomes a Dirac distribution over the reward maximising state (bottom). Thus, minimising expected free energy corresponds to selecting the action such that the predicted states assign most mass to the reward maximising state (bottom-right). $Q^*:=Q(\vec s\mid \vec a^*,s_t)$ denotes the predictive distribution over states given the action sequence that minimises expected free energy $\vec a^*= \arg\min_{\vec a}G(\vec a\mid s_t)$.}
    \label{fig: reaching preferences}
\end{figure}

A proof can be found in Appendix \ref{app: proof 4}. In the zero temperature limit $\beta \to +\infty$, minimising expected free energy corresponds to choosing the action sequence $\vec a$ such that $Q(\vec s\mid \vec a,s_t)$ has most mass on reward maximising states or trajectories (see Figure \ref{fig: reaching preferences}). Of those reward maximising candidates, the minimiser of expected free energy maximises the entropy of future states $\H[Q(\vec s\mid \vec a,s_t)]$, thus leaving options open.

\subsection{Reward maximisation on MDPs with a temporal horizon of $1$}
\label{sec: AI MDP}

In this section, we first consider the case of a single-step decision problem (i.e., a temporal horizon of $T=1$) and demonstrate how the standard active inference scheme maximises reward on this problem in the limit $\beta \to +\infty$. This will act as an important building block for when we subsequently consider more general multi-step decision problems.

The standard decision-making procedure in active inference consists of assigning each action sequence with a probability given by the softmax of the negative expected free energy~\citep{dacostaActiveInferenceDiscrete2020,fristonActiveInferenceProcess2017,barpGeometricMethodsSampling2022}
\begin{align*}
    Q(\vec a \mid s_t) \propto \exp( -G(\vec a \mid s_t)).
\end{align*}

Agents then select the most likely action under this distribution
\begin{align*}
    a_t \in \arg\max_{a \in \A} Q(a\mid s_t) 
     &= \arg\max_{a \in \A} \sum_{\vec a}Q(a\mid \vec a)Q(\vec a \mid s_t) \\
    = \arg\max_{a \in \A} \sum_{\vec a}Q(a\mid \vec a)\exp (-G(\vec a \mid s_t)) &= \arg\max_{a \in \A} \sum_{\substack{\vec a \\\left(\vec a\right)_t = a }}\exp (-G(\vec a \mid s_t)).
\end{align*}

In summary, this scheme selects the first action within action sequences that, on average, maximise their exponentiated negative expected free energies. As a corollary, if the first action is in a sequence with a very low expected free energy, this adds an exponentially large contribution to the selection of this particular action. We summarise this scheme in Table \ref{table: AI on finite horion MDPs}.

\begin{table}[h!]
\centering
\begin{tabular}{c c}
 \hline
 Process & Computation  \\
 \hline
{Perceptual inference}	&  { $Q(\vec s\mid \vec a,s_t)= P(\vec s\mid \vec a,s_t)= \prod_{\tau =t}^{T-1} P(s_{\tau+1}\mid s_{\tau }, a_{\tau})$}  \\
             {Planning as inference}	&  {$G(\vec a\mid s_t)= \operatorname{D_{KL}}[Q(\vec s\mid \vec a,s_t)\mid C(\vec s)]$ }  \\
             Decision-making &  $Q(\vec a\mid s_t)\propto \exp(-G(\vec a\mid s_t))$  \\
             {Action selection}	& $a_t\in \arg\max_{a \in \A} \left[Q(a_t= a\mid s_t)= \sum_{\vec a} Q(a_t=a\mid \vec a)Q(\vec a \mid s_t)\right]$\\
 \hline
\end{tabular}
\caption{Standard active inference scheme on finite horizon MDPs~\citep{barpGeometricMethodsSampling2022}.}
\label{table: AI on finite horion MDPs}
\end{table}

\begin{theorem}
\label{thm: Bellman optimal t1}
In MDPs with known transition probabilities and in the zero temperature limit $\beta \to +\infty$ \eqref{eq: zero temp limit of preferences}, the scheme of Table \ref{table: AI on finite horion MDPs}
\begin{equation}
\label{eq: standard AI state-action policy}
 a_t \in \lim_{\beta \to +\infty} \arg\max_{a \in \A} \sum_{\substack{\vec a \\\left(\vec a\right)_t = a }}\exp (-G(\vec a \mid s_t)), \qquad 
 G(\vec a \mid s_t) = \operatorname{D_{KL}}[Q(\vec s\mid \vec a , s_t)\mid C_\beta (\vec s)]
\end{equation}
is Bellman optimal for the temporal horizon $T=1$.
\end{theorem}

A proof can be found in Appendix \ref{app: proof 5}. Importantly, \emph{the standard active inference scheme \eqref{eq: standard AI state-action policy} falls short in terms of Bellman optimality on planning horizons greater than one}; this rests upon the fact that it does not coincide with backward induction. Recall that backward induction offers a complete description of Bellman optimal state-action policies (Proposition \ref{prop: backward induction construction of Bellman optimal state-action policies}). In contrast, active inference plans by adding weighted expected free energies of each possible future course of action. In other words, unlike backward induction, it considers future courses of action beyond the subset that will subsequently minimise expected free energy, given subsequently encountered states.

\subsection{Reward maximisation on MDPs with finite temporal horizons}
\label{sec: SI MDP}

To achieve Bellman optimality on finite temporal horizons, we turn to the expected free energy of an action given future actions that also minimise expected free energy. To do this we can write the expected free energy recursively, as the immediate expected free energy, plus the expected free energy that one would obtain by subsequently selecting actions that minimise expected free energy~\citep{fristonSophisticatedInference2021}.
The resulting scheme consists of minimising an expected free energy defined recursively, from the last time step to the current timestep. In finite horizon MDPs, this reads
\begin{align*}
        G(a_{T-1}\mid s_{T-1})&= \operatorname{D_{K L}}[Q(s_{T}\mid a_{T-1},s_{T-1})\mid C_\beta(s_{T})] \\
        G(a_\tau\mid s_\tau)&= \operatorname{D_{K L}}[Q(s_{\tau+1}\mid a_\tau,s_\tau)\mid C_\beta(s_{\tau+1})]\\ &+\E_{Q(a_{\tau+1},s_{\tau+1}\mid a_\tau,s_\tau)}[G(a_{\tau+1}\mid s_{\tau+1})], \quad \tau =t,\ldots ,T-2,
\end{align*}
where, at each time-step, actions are chosen to minimise expected free energy
\begin{equation}
\label{eq: future actions that minimise expected free energy}
    Q(a_{\tau+1}\mid s_{\tau+1}) >0 \iff a_{\tau+1} \in  \arg \min_{a\in \A} G(a\mid s_{\tau+1}).
\end{equation}

To make sense of this formulation, we unravel the recursion
\begin{equation}
\label{eq: unpacking recursive EFE}
    \begin{split}
     G(a_t\mid s_t)&= \operatorname{D_{K L}}[Q(s_{t+1}\mid a_t,s_t)\mid C_\beta(s_{t+1})] + \E_{Q(a_{t+1},s_{t+1}\mid a_t,s_t)}[G(a_{t+1}\mid s_{t+1})] \\
    &= \operatorname{D_{K L}}[Q(s_{t+1}\mid a_t,s_t)\mid C_\beta(s_{t+1})] + \E_{Q(a_{t+1},s_{t+1}\mid a_t,s_t)} \left[\operatorname{D_{K L}}[Q(s_{t+2}\mid a_{t+1},s_{t+1})\mid C_\beta(s_{t+2})]\right] \\
     &+ \E_{Q(a_{t+1:t+2},s_{t+1:t+2}\mid a_t,s_t)}[G(a_{t+2}\mid s_{t+2})]\\
     &= \ldots = \E_{Q(\vec a, \vec s\mid a_t,s_t)} \sum_{\tau= t}^{T-1}  \operatorname{D_{K L}}[Q(s_{\tau +1}\mid a_\tau,s_\tau)\mid C_\beta(s_{\tau+1})] \\
     &= \E_{Q(\vec a, \vec s\mid a_t,s_t)}\operatorname{D_{K L}}[Q(\vec s\mid \vec a,s_t)\mid C_\beta(\vec s)],
    \end{split}
\end{equation}
which shows that this expression is exactly the expected free energy under action $a_t$, if one is to pursue future actions that minimise expected free energy \eqref{eq: future actions that minimise expected free energy}. We summarise this 'sophisticated inference' scheme in Table \ref{table: soph AI}.

\begin{table}[h!]
\centering
\begin{tabular}{c c}
 \hline
 Process & Computation  \\
 \hline
             {Perceptual inference}	&  { $Q(s_{\tau+1}\mid a_\tau,s_\tau)= P(s_{\tau+1}\mid a_\tau,s_\tau)$}  \\
             {Planning as inference}	&  $G(a_\tau\mid s_\tau)= \operatorname{D_{K L}}[Q(s_{\tau+1}\mid a_\tau,s_\tau)\mid C_\beta(s_{\tau+1})] \ldots$  \\
             {} & $\qquad \qquad\ldots+\E_{Q(a_{\tau+1},s_{\tau+1}\mid a_\tau,s_\tau)}[G(a_{\tau+1}\mid s_{\tau+1})]$\\
            {Decision-making}	&  $ Q(a_{\tau}\mid s_{\tau}) >0 \iff a_{\tau} \in  \arg \min_{a\in \A} G(a\mid s_{\tau})$  \\
             {Action selection}	& $a_{t} \sim Q(a_{t}\mid s_{t})$
               \\
 \hline
\end{tabular}
\caption{Sophisticated active inference scheme on finite horizon MDPs~\citep{fristonSophisticatedInference2021}.} \label{table: soph AI}
\end{table}

The crucial improvement over the standard active inference scheme (Table \ref{table: AI on finite horion MDPs}) is that planning is now performed based on subsequent counterfactual actions that minimise expected free energy, as opposed to considering all future courses of action. Translating this into the language of state-action policies yields $\forall s\in \S$
\begin{equation}
\label{eq: recursive EFE state action state-action policy}
    \begin{split}
        a_{T-1}(s) &\in \arg\min_{a\in \A} G(a\mid s_{T-1}=s)  \\
        a_{T-2}(s) &\in \arg\min_{a\in \A}  G(a\mid s_{T-2}=s)  \\
        &\vdots \\
        a_{1}(s) &\in \arg\min_{a\in \A}  G(a\mid s_{1}=s)  \\
        a_{0}(s) &\in \arg\min_{a\in \A}  G(a\mid s_{0}). \\
    \end{split}
\end{equation}

Equation \eqref{eq: recursive EFE state action state-action policy} is strikingly similar to the backward induction algorithm (Proposition \ref{prop: backward induction construction of Bellman optimal state-action policies}), and indeed we recover backward induction in the limit $\beta \to +\infty$.

\begin{theorem}[Backward induction as active inference]
\label{thm: backward induction as active inference}
In MDPs with known transition probabilities, and in the zero temperature limit $\beta \to +\infty$ \eqref{eq: zero temp limit of preferences}, the scheme of Table \ref{table: soph AI} 
\begin{equation}
\label{eq: sophisticated AI scheme theorem}
\begin{split}
    Q(a_{\tau}\mid s_{\tau}) >0 &\iff a_{t} \in \lim_{\beta \to +\infty} \arg \min_{a\in \A} G(a\mid s_{\tau}) \\
    G(a_\tau\mid s_\tau)&= \operatorname{D_{K L}}[Q(s_{\tau+1}\mid a_\tau,s_\tau)\mid C_\beta(s_{\tau+1})] +\E_{Q(a_{\tau+1},s_{\tau+1}\mid a_\tau,s_\tau)}[G(a_{\tau+1}\mid s_{\tau+1})]\\  
\end{split}
\end{equation}
is Bellman optimal on any finite temporal horizon as it coincides with the backward induction algorithm from Proposition \ref{prop: backward induction construction of Bellman optimal state-action policies}. Furthermore, if there are multiple actions that maximise future reward, those that are selected by active inference also maximise the entropy of future states $ \H[Q(\vec s\mid \vec a,a,s_0)]$.
\end{theorem}

Note that maximising the entropy of future states keeps the agent's options open~\citep{klyubinKeepYourOptions2008} in the sense of committing the least to a specified sequence of states. A proof of Theorem \ref{thm: backward induction as active inference} can be found in Appendix \ref{app: proof 6}.

\section{Generalisation to POMDPs}
\label{sec: generalisation to POMDPs}

Partially observable Markov decision processes (POMDPs) generalise MDPs in that the agent observes a modality $o_t$, which carries incomplete information about the current state $s_t$, as opposed to the current state itself. 

\begin{definition}[Finite horizon POMDP]
\label{def: POMDP}
A finite horizon POMDP is an MDP (see Definition \ref{def: MDP}) with the following additional data:
\begin{itemize}
    \item $\mathbb O$ a finite set of observations.
    \item \(P(o_t= o\mid s_{t}=s)\) is the probability that the state \(s \in \mathbb S\) at time \(t\) will lead to the observation \(o \in \mathbb O\) at time \(t\). $o_{t}$ are random variables over $\mathbb O$ that correspond to the observation being sampled at time $t=0,...,T$.
\end{itemize}
\end{definition}

\subsection{Active inference on finite horizon POMDPs}

We briefly introduce active inference agents on finite horizon POMDPs with known transition probabilities (for more details, see \citep{dacostaActiveInferenceDiscrete2020,smithStepbystepTutorialActive2022,parrActiveInferenceFree2022}).
We assume that the agent's generative model of its environment is given by the previously defined POMDP (Definition \ref{def: POMDP})\footnote{We do not consider the case where the model parameters have to be learned but comment on it in Appendix \ref{app: reward learning} (details in~\citep{dacostaActiveInferenceDiscrete2020,fristonActiveInferenceLearning2016}).}. 

Let $\vec s:= s_{0:T}, \vec a:= a_{0:T-1}$ be all states and actions (past, present, and future), let $\tilde o:=o_{0:t}$ be the observations available up to time $t$, and $\vec o:=o_{t+1:T}$ be the future observations. The agent has a predictive distribution over states given actions 

\begin{equation*}
     Q(\vec s\mid \vec a,\tilde o):=\prod_{\tau=0}^{T-1} Q(s_{\tau +1}\mid a_\tau, s_{\tau}, \tilde o).
\end{equation*}
that is continuously updated following new observations.

\subsubsection{Perception as inference}
In active inference, perception entails inferences about (past, present, and future) states given observations and a sequence of actions. 
When states are partially observed, the posterior distribution $P(\vec s\mid \vec a,\tilde o)$ is intractable to compute directly. Thus, one approximates it by optimising a variational free energy functional $F_{\vec a}$ (a.k.a. an evidence bound~\citep{bishopPatternRecognitionMachine2006,bealVariationalAlgorithmsApproximate2003,wainwrightGraphicalModelsExponential2007,bleiVariationalInferenceReview2017}) over a space of probability distributions $Q(\cdot\mid \vec a, \tilde o)$ called the \textit{variational family}
\begin{equation}
\label{eq: optim free energy}
\begin{split}
    P(\vec s\mid \vec a,\tilde o)=\arg\min_Q F_{\vec a}[Q(\vec s\mid \vec a,\tilde o)]&= \arg\min_Q \operatorname{D_{KL}}[Q(\vec s\mid \vec a,\tilde o)\mid P(\vec s\mid \vec a,\tilde o)]\\
    F_{\vec a}[Q(\vec s\mid \vec a,\tilde o)] &:= \mathbb E_{Q(\vec s\mid \vec a,\tilde o)}[\log Q(\vec s\mid \vec a,\tilde o) - \log P(\tilde o,\vec s\mid \vec a)].
\end{split}
\end{equation}
Here, $P(\tilde o,\vec s\mid \vec a)$ is the POMDP, which is supplied to the agent, and $P(\vec s\mid \vec a,\tilde o)$. When the free energy minimum \eqref{eq: optim free energy} is reached, the inference is exact 
\begin{equation}
\label{eq: free energy minimum is reached}
   Q(\vec s\mid \vec a,\tilde o)=P(\vec s\mid \vec a,\tilde o).
\end{equation}
For numerical tractability, the variational family may be constrained to a parametric family of distributions, in which case equality is not guaranteed 
\begin{equation}
\label{eq: free energy minimum is almost reached}
    Q(\vec s\mid \vec a,\tilde o)\approx P(\vec s\mid \vec a,\tilde o).
\end{equation}

\subsubsection{Planning as inference}

\begin{figure}[t!]
    \centering
    \includegraphics[width= 400pt]{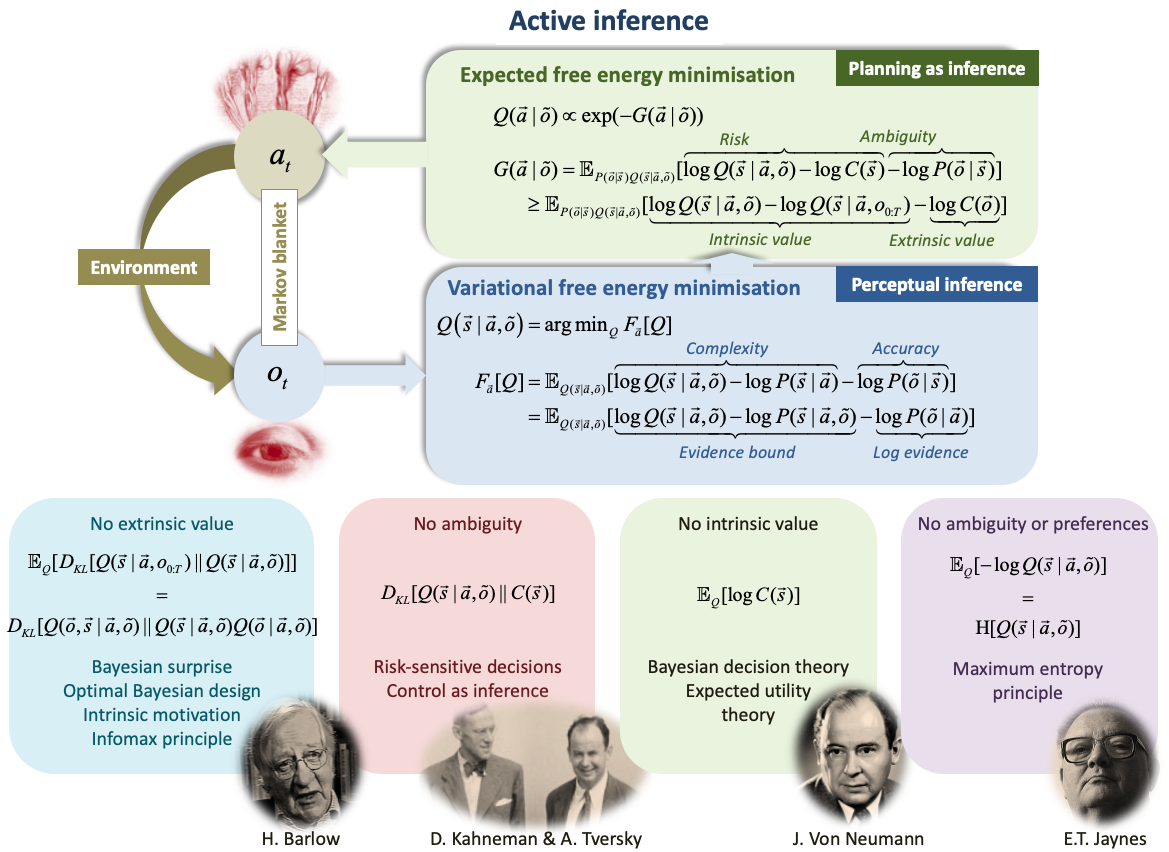}
    \caption{\textbf{Active inference.} The top panels illustrate the perception-action loop in active inference, in terms of minimisation of variational and expected free energy.
    The lower panels illustrate how expected free energy relates to several descriptions of behaviour that predominate in the psychological, machine learning, and economics. These descriptions are disclosed when one removes particular terms from the objective. For example, if we ignore extrinsic value, we are left with intrinsic value, variously known as expected information gain~\citep{mackayInformationTheoryInference2003,lindleyMeasureInformationProvided1956}. This underwrites intrinsic motivation in machine learning and robotics~\citep{bartoNoveltySurprise2013,oudeyerWhatIntrinsicMotivation2007,deciIntrinsicMotivationSelfDetermination1985} and expected Bayesian surprise in visual search~\citep{ittiBayesianSurpriseAttracts2009,sunPlanningBeSurprised2011} and the organisation of our visual apparatus~\citep{linskerPerceptualNeuralOrganization1990,opticanTemporalEncodingTwodimensional1987a,barlowPossiblePrinciplesUnderlying1961,barlowInductiveInferenceCoding1974}. In the absence of ambiguity, we are left with minimising risk, which corresponds to aligning predicted states to preferred states. This leads to risk averse decisions in behavioural economics~\citep{kahnemanProspectTheoryAnalysis1979} and formulations of control as inference in engineering such as KL control ~\citep{vandenbroekRiskSensitivePath2010}. If we then remove intrinsic value, we are left with expected utility in economics~\citep{vonneumannTheoryGamesEconomic1944} that underwrites RL and behavioural psychology~\citep{bartoReinforcementLearningIntroduction1992}. Bayesian formulations of maximising expected utility under uncertainty are also the basis of Bayesian decision theory~\citep{bergerStatisticalDecisionTheory1985}. Finally, if we only consider a fully observed environment with no preferences, minimising expected free energy corresponds to a maximum entropy principle over future states~\citep{jaynesInformationTheoryStatistical1957,jaynesInformationTheoryStatistical1957a}. 
    Note that here $C(o)$ denotes the preferences over observations derived from the preferences over states. These are related by $P(o\mid s)C(s)= P(s\mid o)C(o) $.} 
    \label{fig: EFE}
\end{figure}

The objective that active inference minimises in order the select the best possible courses of action is the \textit{expected free energy}~\citep{barpGeometricMethodsSampling2022,dacostaActiveInferenceDiscrete2020,fristonSophisticatedInference2021}. In POMDPs, the expected free energy reads~\citep{barpGeometricMethodsSampling2022}
\begin{equation*}
\begin{split}
    G(\vec a \mid \tilde o)
    &= \underbrace{\operatorname{D_{KL}}[Q(\vec s\mid \vec a,\tilde o)\mid C_\beta(\vec s)]}_{\text{Risk}}+\underbrace{\E_{Q(\vec s\mid \vec a,\tilde o)}\H
     [P(\vec o\mid \vec s)]}_{\text{Ambiguity}}.\\
\end{split}
\end{equation*}
The expected free energy on POMDPs is the expected free energy on MDPs plus an extra term called ambiguity. This ambiguity term accommodates the uncertainty implicit in partially observed problems. The reason that this resulting functional is called expected free energy is because it comprises a relative entropy (risk) and expected energy (ambiguity). The expected free energy objective subsumes several decision-making objectives that predominate in statistics, machine learning, and psychology, which confers it with several useful properties when simulating behaviour (see Figure \ref{fig: EFE} for details).

\subsection{Maximising reward on POMDPs}

Crucially, our reward maximisation results translate to the POMDP case. To make this explicit, we rehearse Lemma \ref{lemma: reward maxim EFE minim zero temp} in the context of POMDPs.

\begin{proposition}[Reward maximisation on POMDPs]
\label{prop: reward maxim EFE minim zero temp POMDPs}
In POMDPs with known transition probabilities, provided that the free energy minimum is reached \eqref{eq: free energy minimum is reached}, the sequence of actions that minimises expected free energy also maximises expected reward in the zero temperature limit $\beta \to +\infty$ \eqref{eq: zero temp limit of preferences}:
\begin{equation*}
    \lim_{\beta \to +\infty} \arg\min_{\vec a } G(\vec a\mid \tilde o) \subseteq \arg\max_{\vec a} \E_{Q(\vec s\mid \vec a , \tilde o)}[R(\vec s)].
\end{equation*}
Furthermore, of those action sequences that maximise expected reward, the expected free energy minimisers will be those that maximise the entropy of future states minus the (expected) entropy of outcomes given states $\H[Q(\vec s\mid \vec a,\tilde o)] -\E_{Q(\vec s\mid a_t,\tilde o)}\H[P(\vec o\mid \vec s)]]$.
\end{proposition}
From Proposition \ref{prop: reward maxim EFE minim zero temp POMDPs} we see that if there are multiple maximise reward action sequences, those that are selected maximise

\begin{equation*}
    \underbrace{\H[Q(\vec s\mid \vec a,\tilde o)]}_{\text{Entropy of future states}} -\underbrace{\E_{Q(\vec s\mid a_t,\tilde o)}[\operatorname{H
    } [P(\vec o\mid \vec s)]]}_{\text{Entropy of observations given future states}}.
\end{equation*}

In other words, they least commit to a prespecified sequence of future states and ensure that their expected observations are maximally informative of states. Of course, when inferences are inexact, the extent to which Proposition \ref{prop: reward maxim EFE minim zero temp POMDPs} holds depends upon the accuracy of the approximation~\eqref{eq: free energy minimum is almost reached}. A proof of Proposition \ref{prop: reward maxim EFE minim zero temp POMDPs} can be found in Appendix \ref{app: proof 7}. 


The schemes of Table \ref{table: AI on finite horion MDPs} \& \ref{table: soph AI} exist in the POMDP setting, (e.g.,~\citep{barpGeometricMethodsSampling2022} and~\citep{fristonSophisticatedInference2021}, respectively). Thus, in POMDPs with known transition probabilities, provided that inferences are exact \eqref{eq: free energy minimum is reached} and in the zero temperature limit $\beta \to +\infty$ \eqref{eq: zero temp limit of preferences}, standard active inference~\citep{barpGeometricMethodsSampling2022} maximises reward on temporal horizons of $1$ but not beyond, and a recursive scheme such as sophisticated active inference~\citep{fristonSophisticatedInference2021} maximises reward on finite temporal horizons. Note that, for computational tractability, the sophisticated active inference scheme presented in~\citep{fristonSophisticatedInference2021} does not generally perform exact inference; thus, the extent to which it will maximise reward in practice will depend upon the accuracy of its inferences. Nevertheless, our results indicate that sophisticated active inference will vastly outperform standard active inference in most reward maximisation tasks.

\section{Discussion}
\label{sec: discussion}


In this paper, we have examined a specific notion of optimality; namely, Bellman optimality; defined as selecting actions to maximise future expected rewards. We demonstrated how and when active inference is Bellman optimal on finite horizon POMDPs with known transition probabilities and reward function. 

These results highlight important relationships between active inference, stochastic control, and RL, as well as conditions under which they would and would not be expected to behave similarly (e.g., environments with multiple reward-maximising trajectories, those affording ambiguous observations, etc.). We refer the reader to Appendix \ref{app: AIF and RL} for a broader discussion of the relationship between active inference and reinforcement learning.

\subsection{Decision-making beyond reward maximisation}
More broadly, it is important to ask if reward maximisation is the right objective underwriting intelligent decision-making? This is an important question for decision neuroscience. That is, do humans optimise a reward signal, expected free energy, or other planning objectives. This can be addressed by comparing the evidence for these competing hypotheses based on empirical data (e.g., see~\citep{smithImpreciseActionSelection2020a,smithSlowerLearningRates2022,smithGreaterDecisionUncertainty2021,smithLongtermStabilityComputational2021}). Current empirical evidence suggests that humans are not purely reward-maximising agents: they also engage in both random and directed exploration~\citep{dawCorticalSubstratesExploratory2006,mirzaHumanVisualExploration2018,wilsonHumansUseDirected2014,gershmanDeconstructingHumanAlgorithms2018,schulzAlgorithmicArchitectureExploration2019,wilsonBalancingExplorationExploitation2021,xu2021novelty} and keep their options open~\citep{schwartenbeckEvidenceSurpriseMinimization2015}. As we have illustrated, active inference implements a clear form of directed exploration through minimising expected free energy. Although not covered in detail here, active inference can also accommodate random exploration by sampling actions from the posterior belief over action sequences, as opposed to selecting the most likely action as presented in Tables \ref{table: AI on finite horion MDPs} and \ref{table: soph AI}.

Note that behavioural evidence favouring models that do not solely maximise reward within reward maximisation tasks---i.e., where "maximise reward" is the explicit instruction---is not a contradiction. Rather, gathering information about the environment (exploration) generally helps to reap more reward in the long run, as opposed to greedily maximising reward based on imperfect knowledge~\citep{sajidActiveInferenceDemystified2021,cullenActiveInferenceOpenAI2018}. This observation is not new and many approaches to simulating adaptive agents employed today differ significantly from their reward maximising antecedents (Appendix \ref{app: exploration exploitation dilemma}).

\subsection{Learning}

When the transition probabilities or reward function are unknown to the agent, the problem becomes one of reinforcement learning (RL)~\citep{shohamMultiagentReinforcementLearning2003} as opposed to stochastic control. Although we did not explicitly consider it above, this scenario can be accommodated by active inference by simply equipping the generative model with a prior, and updating the model via variational Bayesian inference to best fit observed data. Depending on the specific learning problem and generative model structure, this can involve updating the transition probabilities and/or the target distribution $C$. In POMDPs it can also involve updating the probabilities of observations under each state. We refer to Appendix \ref{app: reward learning} for discussion of reward learning through active inference and connections to representative RL approaches, and~\citep{dacostaActiveInferenceDiscrete2020,fristonActiveInferenceLearning2016} for learning transition probabilities through active inference.

\subsection{Scaling active inference}

When comparing RL and active inference approaches generally, one outstanding issue for active inference is whether it can be scaled up to solve the more complex problems currently handled by RL in machine learning contexts~\citep{catalLearningPerceptionPlanning2020a,tschantzScalingActiveInference2019,millidgeDeepActiveInference2020,catalRobotNavigationHierarchical2021,fountasDeepActiveInference2020,mazzagliaContrastiveActiveInference2021}. This is an area of active research.

One important issue along these lines is that planning ahead by evaluating all or many possible sequences of actions is computationally prohibitive in many applications. Three complementary solutions that have emerged are: 1) employing hierarchical generative models that factorise decisions into multiple levels and reduce the size of the decision tree by orders of magnitude~\citep{fristonDeepTemporalModels2018,catalRobotNavigationHierarchical2021,parrComputationalNeurologyMovement2021}, 2) efficiently searching the decision tree using algorithms like Monte Carlo tree search~\citep{silverMasteringGameGo2016,fountasDeepActiveInference2020,maistoActiveTreeSearch2021,championBranchingTimeActive2021,championBranchingTimeActive2021a}, and 3) amortising planning using artificial neural networks~\citep{catalLearningPerceptionPlanning2020a,fountasDeepActiveInference2020,sajidExplorationPreferenceSatisfaction2021}. 

Another issue rests upon learning the generative model. Active inference may readily learn the parameters of a generative model; however, more work needs to be done on devising algorithms for learning the structure of generative models themselves~\citep{smithActiveInferenceApproach2020,fristonActiveInferenceCuriosity2017}. This is an important research problem in generative modelling, called Bayesian model selection or structure learning~\citep{gershmanLearningLatentStructure2010,tervoNeuralImplementationStructure2016}.

Note that these issues are not unique to active inference. Model-based RL algorithms deal with the same combinatorial explosion when evaluating decision trees, which is one primary motivation for developing efficient model-free RL algorithms. However, other heuristics have also been developed for efficiently searching and pruning decision trees in model-based RL, e.g.,~\citep{huysBonsaiTreesYour2012,lallyNeuralBasisAversive2017}. Furthermore, model-based RL suffers the same limitation regarding learning generative model structure. Yet, RL may have much to offer active inference in terms of efficient implementation and the identification of methods to scale to more complex applications~\citep{mazzagliaContrastiveActiveInference2021,fountasDeepActiveInference2020}.

\section{Conclusion}
\label{sec: conclusion}

In summary, we have shown that under the specification that the active inference agent prefers maximising reward \eqref{eq: zero temp limit of preferences}:
\begin{enumerate}
    \item On finite horizon POMDPs with known transition probabilities, the objective optimised for action selection in active inference (i.e., expected free energy) produces reward maximising action sequences when state-estimation is \textit{exact}. When there are multiple reward maximising candidates, this selects those sequences that maximise the entropy of future states---thereby keeping options open---and that minimise the ambiguity of future observations so that they are are maximally informative. More generally, the extent to which action sequences will be reward maximising will depend on the accuracy of state-estimation.  
    \item The standard active inference scheme (e.g., ~\citep{barpGeometricMethodsSampling2022}) produces Bellman optimal actions for planning horizons of $1$ when state-estimation is exact, but not beyond.
    \item A sophisticated active inference scheme (e.g., \citep{fristonSophisticatedInference2021}) produces Bellman optimal actions on any finite planning horizon when state-estimation is \textit{exact}. Furthermore, this scheme generalises the well-known backward induction algorithm from dynamic programming to partially observed environments. Note that, for computational efficiency, the scheme presented in \citep{fristonSophisticatedInference2021} does not generally perform exact state-estimation; thus, the extent to which it will maximise reward in practice will depend upon the accuracy of its inferences. Nevertheless, it is clear from our results that sophisticated active inference will vastly outperform standard active inference in most reward maximisation tasks.
\end{enumerate}

Note that, for computational tractability, the sophisticated active inference scheme presented in~\citep{fristonSophisticatedInference2021} does not generally perform exact inference; thus, the extent to which it will maximise reward in practice will depend upon the accuracy of its inferences. Nevertheless, it is clear from these results that sophisticated active inference will vastly outperform standard active inference in most reward maximisation tasks.

\section*{Acknowledgements}
The authors thank Dimitrije Markovic and Quentin Huys for providing helpful feedback during the preparation of the manuscript.

\section*{Funding information}
LD is supported by the Fonds National de la Recherche, Luxembourg (Project code: 13568875). NS is funded by the Medical Research Council (MR/S502522/1) and 2021-2022 Microsoft PhD Fellowship. KF is supported by funding for the Wellcome Centre for Human Neuroimaging (Ref: 205103/Z/16/Z), a Canada-UK Artificial Intelligence Initiative (Ref: ES/T01279X/1) and the European Union’s Horizon 2020 Framework Programme for Research and Innovation under the Specific Grant Agreement No. 945539 (Human Brain Project SGA3). RS is supported by the William K. Warren Foundation, the Well-Being for Planet Earth Foundation, the National Institute for General Medical sciences (P20GM121312), and the National Institute of Mental Health (R01MH123691). This publication is based on work partially supported by the EPSRC Centre for Doctoral Training in Mathematics of Random Systems: Analysis, Modelling and Simulation (EP/S023925/1).

\section*{Author contributions}
LD: conceptualisation, proofs, writing -- first draft, review and editing. NS, TP, KF, RS: conceptualisation, writing -- review and editing.

\bibliography{bib}

\appendix

\section{Active inference and reinforcement learning}
\label{app: AIF and RL}

This paper considered how active inference can solve the stochastic control problem. 
In this appendix, we discuss the broader relationship between active inference and RL.

Loosely speaking, RL is the field of methodologies and algorithms that learn reward-maximising actions from data and seek to maximise reward in the long run. Because RL is a data-driven field, algorithms are selected based on how well they perform on benchmark problems. This has produced a plethora of diverse algorithms, many designed to solve specific problems, each with their own strengths and limitations. This makes RL difficult to characterise as a whole. Thankfully, many approaches to model-based RL and control can be traced back to approximating the optimal solution to the Bellman equation~\citep{bellmanAppliedDynamicProgramming2015,bertsekasStochasticOptimalControl1996} (although this may become computationally intractable in high-dimensions~\citep{bartoReinforcementLearningIntroduction1992}). Our results showed how and when decisions under active inference and such RL approaches are similar. 

This appendix discusses how active inference and RL relate and differ more generally. Their relationship has become increasingly important to understand, as a growing body of research has begun to 1) compare the performance of active inference and RL models in simulated environments~\citep{sajidActiveInferenceDemystified2021,cullenActiveInferenceOpenAI2018,millidgeDeepActiveInference2020}, 2) apply active inference to model human behaviour on reward learning tasks~\citep{smithSlowerLearningRates2022,smithGreaterDecisionUncertainty2021,smithLongtermStabilityComputational2021,smithImpreciseActionSelection2020a}, and 3) consider the complementary predictions and interpretations they each offer in computational neuroscience, psychology, and psychiatry~\citep{schwartenbeckComputationalMechanismsCuriosity2019,schwartenbeckDopaminergicMidbrainEncodes2015,tschantzLearningActionorientedModels2020,cullenActiveInferenceOpenAI2018,huysBonsaiTreesYour2012}.

\subsection{Main differences between active inference and reinforcement learning}



\paragraph{Philosophy.} 
Active inference and RL differ profoundly in their philosophy. RL derives from the normative principle of maximising reward~\citep{bartoReinforcementLearningIntroduction1992}, while active inference describes systems that maintain their structural integrity over time~\citep{fristonFreeEnergyPrinciple2022,barpGeometricMethodsSampling2022}. Despite this difference, there are many practical similarities between these frameworks. For example, recall that behaviour in active inference is completely determined by the agent's preferences, determined as priors in their generative model. Crucially, log priors can be interpreted as reward functions and vice-versa, which is how behaviour under RL and active inference can be related.

\paragraph{Model based and model free.} Active inference agents \textit{always} embody a generative (i.e., forward) model of their environment, while RL comprises both model-based algorithms and simpler model-free algorithms. In brief, 'model-free' means that agents learn a reward-maximising state-action mapping, based on updating cached state-action pair values, through initially random actions that do not consider future state transitions. In contrast, model-based RL algorithms attempt to extend stochastic control approaches by learning the dynamics and reward function from data. Recall that stochastic control calls on strategies that evaluate different actions on a carefully handcrafted forward model of dynamics (i.e., known transition probabilities) to finally execute the reward-maximising action. Under this terminology, \textit{all} active inference agents are model-based.

\paragraph{Modelling exploration.}
Exploratory behaviour---which can improve reward maximisation in the long run---is implemented differently in the two approaches. In most cases, RL implements a simple form of exploration by incorporating randomness in decision-making~\citep{tokicValueDifferenceBasedExploration2011,wilsonHumansUseDirected2014}, where the level of randomness may or may not change over time as a function of uncertainty. In other cases, RL incorporates ad-hoc information bonuses in the reward function or other decision-making objectives to build in directed exploratory drives (e.g., upper confidence bound algorithms or Thompson sampling). 
In contrast, directed exploration emerges naturally within active inference through interactions between the risk and ambiguity terms in the expected free energy~\citep{schwartenbeckComputationalMechanismsCuriosity2019,dacostaActiveInferenceDiscrete2020}. This addresses the explore-exploit dilemma and confers the agent with artificial curiosity~\citep{fristonActiveInferenceCuriosity2017,schwartenbeckComputationalMechanismsCuriosity2019,schmidhuberFormalTheoryCreativity2010,stillInformationtheoreticApproachCuriositydriven2012}, as opposed to the need to add ad-hoc information bonus terms~\citep{tokicValueDifferenceBasedExploration2011}. We expand on this relationship further in Appendix \ref{app: exploration exploitation dilemma}.

\paragraph{Control and learning as inference.}
Active inference integrates state-estimation, learning, decision-making, and motor control under the single objective of minimising free energy~\citep{dacostaActiveInferenceDiscrete2020}. In fact, active inference extends previous work on the duality between inference and control~\citep{todorovGeneralDualityOptimal2008,kappenOptimalControlGraphical2012,rawlikStochasticOptimalControl2013,toussaintRobotTrajectoryOptimization2009} to solve motor control problems via approximate inference (i.e., planning as inference)~\citep{fristonActiveInferenceAgency2012,millidgeRelationshipActiveInference2020a,fristonReinforcementLearningActive2009, attiasPlanningProbabilisticInference2003,botvinickPlanningInference2012}. Therefore, some of the closest RL methods to active inference are control as inference, also known as maximum entropy RL~\citep{levineReinforcementLearningControl2018, millidgeRelationshipActiveInference2020a,ziebartModelingPurposefulAdaptive2010}, though one major difference is in the choice of decision-making objective. Loosely speaking, these aforementioned methods minimise the risk term of the expected free energy, while active inference also minimises ambiguity.

\paragraph{Useful features of active inference.}
\begin{enumerate}[wide, labelwidth=!, labelindent=0.2pt]
    \item Active inference allows great flexibility and transparency when modelling behaviour. It affords explainable decision-making as a mixture of information- and reward-seeking policies that are explicitly encoded (and evaluated in terms of expected free energy) in the generative model as priors, which are specified by the user~\citep{dacostaHowActiveInference2022a}. As we have seen, the kind of behaviour that can be produced includes the optimal solution to the Bellman equation.
    \item Active inference accommodates deep hierarchical generative models combining both discrete and continuous state-spaces~\citep{fristonDeepTemporalModels2018,fristonGraphicalBrainBelief2017,parrComputationalNeurologyMovement2021}.
    \item The expected free energy objective optimised during planning subsumes many approaches used to describe and simulate decision-making in the physical, engineering, and life sciences, affording it various interesting properties as an objective (Figure \ref{fig: EFE} and~\citep{fristonSophisticatedInference2021}). For example, exploratory and exploitative behaviour are canonically integrated, which finesses the need for manually incorporating ad-hoc exploration bonuses in the reward function~\citep{dacostaActiveInferenceModel2022}.
    \item Active inference goes beyond state-action policies that predominate in traditional RL to sequential policy optimisation. In sequential policy optimisation, one relaxes the assumption that the same action is optimal given a particular state---and acknowledges that the sequential order of actions may matter. This is similar to the linearly-solvable MDP formulation presented by~\citep{todorov2007linearly,todorov2009efficient}, where transition probabilities directly determine actions, and an optimal policy specifies transitions that minimise some divergence cost. This way of approaching policies is perhaps most apparent in terms of exploration. Put simply, it is clearly better to explore and then exploit than the converse. Because expected free energy is a functional of beliefs, exploration becomes an integral part of decision-making---in contrast with traditional RL approaches that try to optimise a reward function of states. In other words, active inference agents will explore until enough uncertainty is resolved for reward maximising, goal-seeking imperatives to start to predominate.
\end{enumerate}
Such advantages should motivate future research to better characterise the environments in which these properties offer useful advantages---such as where performance benefits from learning and planning at multiple temporal scales, and from the ability to select policies that resolve both state and parameter uncertainty.

\subsection{Reward learning}
\label{app: reward learning}

Given the focus on relating active inference to the objective of maximising reward, it is worth briefly illustrating how active inference can learn the reward function from data and its potential connections to representative RL approaches. One common approach for active inference to learn a reward function~\citep{smithImpreciseActionSelection2020a,smithSlowerLearningRates2022} is to set preferences over observations rather than states, which corresponds to assuming that inferences over states given outcomes are accurate
\begin{equation*}
\begin{split}
    \underbrace{\operatorname{D_{KL}}\left[Q\left(\vec s\mid  \vec a, \tilde o\right) \mid  C\left(\vec s\right)\right]}_{\text {Risk }(\text {states})} &=\underbrace{\operatorname{D_{KL}}\left[Q\left(\vec o \mid  \vec a, \tilde o\right) \mid  C\left(\vec o\right)\right]}_{\text {Risk (outcomes)}}+\underbrace{\mathbb{E}_{Q\left(\vec o \mid  \vec a, \tilde o\right)}\left[\operatorname{D_{KL}}\left[Q\left(\vec s \mid  \vec o,\tilde o, \vec a\right) \mid  P\left(\vec s \mid  \vec o\right)\right]\right]}_{\approx 0} \\
    & \approx \underbrace{\operatorname{D_{KL}}\left[Q\left(\vec o \mid  \vec a, \tilde o\right) \mid  C\left(\vec o\right)\right]}_{\text {Risk (outcomes)}},
\end{split}
\end{equation*}
i.e., equality holds whenever the free energy minimum is reached \eqref{eq: free energy minimum is reached}. Then one sets the preference distribution such that the observations designated as rewards are most preferred. In the zero temperature limit \eqref{eq: zero temp limit of preferences}, preferences only assign mass to reward-maximising observations.
When formulated in this way, the reward signal is treated as sensory data, as opposed to a separate signal from the environment. When one sets allowable actions (controllable state transitions) to be fully deterministic such that the selection of each action will transition the agent to a given state with certainty, the emerging dynamics are such that the agent chooses actions to resolve uncertainty about the probability of observing reward under each state. Thus, learning the reward probabilities of available actions amounts to learning the likelihood matrix $P(\vec o\mid \vec s) := o_t \cdot As_t$, where $A$ is a stochastic matrix. This is done by setting a prior $\mathbf a$ over $A$, i.e., a matrix of non-negative components, the columns of which are Dirichlet priors over the columns of $A$. The agent then learns by accumulating Dirichlet parameters. Explicitly, at the end of a trial or episode, one sets~\citep{dacostaActiveInferenceDiscrete2020,fristonActiveInferenceLearning2016}
\begin{align}
\label{eq: accumulate Dirichlet parameters}
    \mathbf{a}\leftarrow \mathbf a+\sum_{\tau=0}^{T} o_{\tau} \otimes Q(s_\tau\mid o_{0:T})
\end{align}
In \eqref{eq: accumulate Dirichlet parameters}, $Q(s_\tau\mid o_{0:T})$ is seen as a vector of probabilities over the state-space $\S$, corresponding 
to the probability of having been in one or another state at time the $\tau$ after having gathered observations throughout the trial. This rule simply amounts to counting observed state-outcome pairs, which is equivalent to state-reward pairs when the observation modalities correspond to reward.

One should not conflate this approach with the update rule consisting of accumulating state-observation counts in the likelihood matrix
\begin{align}
\label{eq: naive update rule}
    A\leftarrow A+\sum_{\tau=0}^{T} o_{\tau} \otimes Q(s_\tau\mid o_{0:T})
\end{align}
and then normalising its columns to sum to one when computing probabilities. The latter simply approximates the likelihood matrix $A$ by accumulating the number of observed state-outcome pairs. This is distinct from the approach outlined above, which encodes uncertainty over the matrix $A$, as a probability distribution over possible distributions $P(o_t\mid s_t)$. The agent is initially very unconfident about $A$, which means that it doesn't place high probability mass on any specification of $P(o_t\mid s_t)$. This uncertainty is gradually resolved by observing state-observation (or state-reward) pairs. Computationally, it is a general fact of Dirichlet priors that an increase in elements of $\mathbf a$ causes the entropy of $P(o_t\mid s_t)$ to decrease. As the terms added in \eqref{eq: accumulate Dirichlet parameters} are always positive, one choice of distribution $P(o_t\mid s_t)$---which best matches available data and prior beliefs---is ultimately singled out. In other words, the likelihood mapping is learned.

The update rule consisting of accumulating state-observation counts in the likelihood matrix \eqref{eq: naive update rule} (i.e., not incorporating Dirichlet priors) bears some similarity to off-policy learning algorithms such as Q-learning. In Q-learning, the objective is to find the best action given the current observed state. For this, the Q-learning agent accumulates values for state-action pairs with repeated observation of rewarding/punishing action outcomes---much like state-observation counts. This allows it to learn the Q-value function that defines a reward maximising policy. 

As always in partially observed environments, we cannot guarantee that the true likelihood mapping will be learned in practice. Please see~\citep{smithActiveInferenceModel2019} for examples where, although not in an explicit reward-learning context, learning the likelihood can be more or less successful in different situations. Learning the true likelihood fails when the inference over states is inaccurate, such as when using too severe of a mean-field approximation to the free energy~\citep{bleiVariationalInferenceReview2017,parrNeuronalMessagePassing2019,tanakaTheoryMeanField1999}, which causes the agent to misinfer states and thereby accumulate Dirichlet parameters in the wrong locations. Intuitively, this amounts to jumping to conclusions too quickly.

\begin{remark}
If so desired, reward learning in active inference can also be equivalently formulated as learning transition probabilities $P(s_{t+1}\mid s_t,a_t)$. In this alternative setup (as exemplified in~\citep{salesLocusCoeruleusTracking2019}), mappings between reward states and reward outcomes in $A$ are set as identity matrices, and the agent instead learns the probability of transitioning to states that deterministically generate preferred (rewarding) observations given the choice of each action sequence. The transition probabilities under each action are learned in a similar fashion as above \eqref{eq: accumulate Dirichlet parameters}, by accumulating counts on a Dirichlet prior over $P(s_{t+1}\mid s_{t},a_t)$. See \cite[Appendix]{dacostaActiveInferenceDiscrete2020} for details.
\end{remark}

Given the model-based Bayesian formulation of active inference, more direct links can be made between the active inference approach to reward learning described above and other Bayesian model-based RL approaches. For such links to be realised, the Bayesian RL agent would be required to have a prior over a prior (e.g., a prior over the reward function prior or transition function prior). One way to implicitly incorporate this is through Thompson sampling~\citep{ghavamzadeh2016bayesian,russo2017tutorial,russo2014learning,russo2016information}. While not the focus of this paper, future work could further examine the links between reward learning in active inference and model-based Bayesian RL schemes.

\subsection{Solving the exploration-exploitation dilemma}
\label{app: exploration exploitation dilemma}

An important distinction between active inference and reinforcement learning schemes is how they solve the exploration-exploitation dilemma.

The exploration-exploitation dilemma~\citep{berger-talExplorationExploitationDilemmaMultidisciplinary2014} arises whenever an agent has incomplete information about its environment, such as when the environment is partially observed, or the generative model has to be learned. The dilemma is then about deciding whether to execute actions aiming to collect reward based on imperfect information about the environment, or to execute actions aiming to gather more information---allowing the agent to reap more reward in the future. Intuitively, it is always best to explore and then exploit, but optimising this trade-off can be difficult.

Active inference balances exploration and exploitation through minimising the risk and ambiguity inherent in the minimisation of expected free energy. This balance is context-sensitive and can be adjusted by modifying the agent's preferences~\citep{dacostaHowActiveInference2022a}. In turn, the expected free energy is obtained from a description of agency in biological systems derived from physics ~\citep{barpGeometricMethodsSampling2022,fristonFreeEnergyPrinciple2022}. 

Modern RL algorithms integrate exploratory and exploitative behaviour in many different ways. One option is curiosity-driven rewards to encourage exploration. Maximum entropy RL and control-as-inference make decisions by minimising a KL divergence to the target distribution~\citep{todorovGeneralDualityOptimal2008,ziebart2008maximum,haarnoja2017reinforcement,levine2018reinforcement, sac,eysenbach2019if}, which combines reward maximisation with maximum entropy over states. This is similar to active inference on MDPs~\citep{millidgeRelationshipActiveInference2020a}. Similarly, the model-free Soft Actor-Critic~\citep{sac} algorithm maximises both expected reward and entropy. This outperforms other state-of-the-art algorithms in continuous control environments and has been shown to be more sample efficient than its reward-maximising counterparts~\citep{sac}. Hyper~\cite{zintgraf2021exploration} proposes reward maximisation alongside minimising uncertainty over both external states and model parameters. Bayes-adaptive RL~\citep{ rossBayesAdaptivePOMDPs2008,rossBayesianApproachLearning,guezEfficientBayesAdaptiveReinforcement2013, guezScalableEfficientBayesAdaptive2013,zintgrafVariBADVeryGood2020} provides policies that balance exploration and exploitation with the aim of maximising reward. Thompson sampling provides a way to balance exploiting current knowledge to maximise immediate performance and accumulating new information to improve future performance~\citep{russo2017tutorial}. This reduces to optimising dual objectives, reward maximisation and information gain, similar to active inference on POMDPs.
Empirically, \citet{sajidActiveInferenceDemystified2021} demonstrated that an active inference agent and a Bayesian model-based RL agent using Thompson sampling exhibit similar behaviour when preferences are defined over outcomes. They also highlighted that, when completely removing the reward signal from the environment, the two agents both select policies that maximise some sort of information gain.

In general, the way each of these approaches to the exploration-exploitation dilemma differ in theory and in practice remains largely unexplored.

\section{Proofs}

\subsection{Proof of Proposition \ref{prop: Existence of the Bellman optimal state-action policy}}
\label{app: proof 1}

Note that a Bellman optimal state-action policy $\Pi^*$ is a maximal element according to the partial ordering $\leq$. Existence thus consists of a simple application of Zorn's lemma. Zorn's lemma states that if any increasing chain
\begin{equation}
\label{eq: chain of state-action policies}
    \Pi_1 \leq \Pi_2\leq \Pi_3 \leq \ldots
\end{equation}
has an upper bound that is a state-action policy, then there is a maximal element $\Pi^*$.

Given the chain \eqref{eq: chain of state-action policies}, we construct an upper bound. We enumerate $ \mathbb A \times \mathbb S \times \mathbb T$ by $(\alpha_1,\sigma_1,t_1),\ldots,$\linebreak$(\alpha_N,\sigma_N,t_N)$. Then the state-action policy sequence
\begin{equation*}
    \Pi_n (\alpha_1\mid \sigma_1,t_1), \quad n=1,2,3,\ldots
\end{equation*}
 is bounded within $[0,1]$. By the Bolzano-Weierstrass theorem, there exists a subsequence $\Pi_{n_k} (\alpha_1\mid \sigma_1,t_1)$, $ k=1,2,3, \ldots$ that converges. Similarly, $\Pi_{n_k} (\alpha_2\mid \sigma_2,t_2)$ is also a bounded sequence, and by Bolzano-Weierstrass it has a subsequence $\Pi_{n_{k_j}} (a_2\mid \sigma_2,t_2)$ that converges. We repeatedly take subsequences until $N$. To ease notation, call the resulting subsequence $\Pi_m$, $m=1,2,3, \ldots$

With this, we define $\hat \Pi= \lim_{m \to \infty }\Pi_m $. It is straightforward to see that $\hat \Pi$ is a state-action policy:
\begin{equation*}
\begin{split}
    \hat\Pi(\alpha\mid \sigma, t) &= \lim_{m \to \infty} \Pi_m(\alpha\mid \sigma, t) \in [0,1], \quad \forall (\alpha, \sigma, t) \in \mathbb A\times \mathbb S\times \mathbb T,\\
    \sum_{\alpha \in\mathbb  A} \hat\Pi(\alpha\mid \sigma, t) &=\lim_{m \to \infty} \sum_{\alpha \in \mathbb A} \Pi_m(\alpha\mid \sigma, t) =1, \quad \forall (\sigma, t) \in \mathbb S\times \mathbb T.
\end{split}
\end{equation*}

To show that $\hat \Pi$ is an upper bound, take any $\Pi$ in the original chain of state-action policies \eqref{eq: chain of state-action policies}. Then by the definition of an increasing subsequence, there exists an index $M \in \mathbb N$ such that $\forall k \geq M$: $\Pi_k \geq \Pi$. Since limits commute with finite sums, we have $v_{\hat \Pi}(s,t)=\lim_{m \to \infty }v_{\Pi_m}(s,t) \geq v_{\Pi_k}(s,t)  \geq v_{\Pi}(s,t) $ for any $(s,t) \in \mathbb S \times \mathbb T$. Thus, by Zorn's lemma there exists a Bellman optimal state-action policy $\Pi^*$.

\subsection{Proof of Proposition \ref{prop: charac Bellman optimal state-action policy}}
\label{app: proof 2}

$1) \Rightarrow 2):$ We only need to show assertion \textit{(b)}. By contradiction, suppose that $\exists (s,\alpha) \in \S\times \A$ such that $\Pi(\alpha\mid s,0)>0$ and 
\begin{equation*}
  \E_{\Pi}[R(s_{1:T})\mid s_0=s,a_0=\alpha]
  < \max_{a \in \A} \E_{\Pi}[R(s_{1:T})\mid s_0=s,a_0=a].
\end{equation*}
We let $\alpha'$ be the Bellman optimal action at state $s$ and time $0$ defined as
\begin{equation*}
    \alpha' :=\arg \max_{a\in \A} \E_{\Pi}[R(s_{1:T})\mid s_0=s,a_0=a].
\end{equation*}
Then, we let $\Pi'$ be the same state-action policy as $\Pi$ except that $\Pi'(\cdot\mid s,0)$ assigns $\alpha'$ deterministically. Then,

\begin{equation*}
    \begin{split}
        v_{\Pi}(s, 0)&= \sum_{a \in \A} \E_{\Pi}[R(s_{1:T})\mid s_0 =s,a_0=a] \Pi(a\mid s, 0)\\
        &< \max_{a\in \mathbb A} \E_{\Pi}[R(s_{1:T})\mid s_0=s,a_0=a]\\
        &= \E_{\Pi'}[R(s_{1:T})\mid s_0=s,a_0=\alpha'] \Pi'(\alpha'\mid s,0)\\
        &= \sum_{a \in \A} \E_{\Pi'}[R(s_{1:T})\mid s_0 =s,a_0=a] \Pi'(a\mid s, 0) \\
        &=v_{\Pi'}(s, 0).
    \end{split}
\end{equation*}
So $\Pi$ is not Bellman optimal, which is a contradiction.

$1) \Leftarrow 2):$ We only need to show that $\Pi$ maximises $v_\Pi(s,0), \forall s\in \S$. By contradiction, there exists a state-action policy $\Pi'$ and a state $s\in \S$ such that
\begin{equation*}
\begin{split}
    v_\Pi(s,0) &< v_{\Pi'}(s,0) \\
    \iff \sum_{a \in \A} \E_{\Pi}[R(s_{1:T})\mid s_0 =s,a_0=a] \Pi(a\mid s, 0) &< \sum_{a \in \A} \E_{\Pi'}[R(s_{1:T})\mid s_0 =s,a_0=a] \Pi'(a\mid s, 0).
\end{split}
\end{equation*}

By \textit{(a)} the left hand side equals 
\begin{equation*}
    \max_{a\in \A} \E_{\Pi}[R(s_{1:T})\mid s_0=s,a_0=a].
\end{equation*}

Unpacking the expression on the right-hand side:

\begin{equation}
\label{eq: unpacking 1<-2}
\begin{split}
    &\sum_{a \in \A} \E_{\Pi'}[R(s_{1:T})\mid s_0 =s,a_0=a] \Pi'(a\mid s, 0)\\
    &= \sum_{a \in \A} \sum_{\sigma \in \S} \E_{\Pi'}[R(s_{1:T})\mid s_1 =\sigma]P(s_1=\sigma\mid s_0=s , a_0=a)\Pi'(a\mid s, 0) \\
    &= \sum_{a \in \A} \sum_{\sigma \in \S} \left\{ \E_{\Pi'}[R(s_{2:T})\mid s_1 =\sigma] +R(\sigma) \right\}P(s_1=\sigma\mid s_0=s , a_0=a)\Pi'(a\mid s, 0) \\
    &=\sum_{a \in \A} \sum_{\sigma \in \S} \left\{ v_{\Pi'}(\sigma,1) +R(\sigma)\right] P(s_1=\sigma\mid s_0=s , a_0=a)\Pi'(a\mid s, 0)
\end{split}
\end{equation}
Since $\Pi$ is Bellman optimal when restricted to $\{1,\ldots,T\}$ we have $v_{\Pi'}(\sigma, 1) \leq v_{\Pi}(\sigma, 1), \forall \sigma \in \S$. Therefore,
\begin{equation*}
\begin{split}
    & \sum_{a \in \A} \sum_{\sigma \in \S} \left\{ v_{\Pi'}(\sigma,1) +R(\sigma)\right] P(s_1=\sigma\mid s_0=s , a_0=a)\Pi'(a\mid s, 0) \\
    &\leq \sum_{a \in \A} \sum_{\sigma \in \S} \left\{ v_{\Pi}(\sigma,1) +R(\sigma)\right] P(s_1=\sigma\mid s_0=s , a_0=a)\Pi'(a\mid s, 0).
\end{split}
\end{equation*}
Repeating the steps above \eqref{eq: unpacking 1<-2}, but in reverse order, yields

\begin{equation*}
    \sum_{a \in \A} \E_{\Pi'}[R(s_{1:T})\mid s_0 =s,a_0=a] \Pi'(a\mid s, 0) \leq \sum_{a \in \A} \E_{\Pi}[R(s_{1:T})\mid s_0 =s,a_0=a] \Pi'(a\mid s, 0)
\end{equation*}

However,
\begin{equation*}
    \sum_{a \in \A} \E_{\Pi}[R(s_{1:T})\mid s_0 =s,a_0=a] \Pi'(a\mid s, 0) < \max_{a\in \A} \E_{\Pi}[R(s_{1:T})\mid s_0=s,a_0=a]
\end{equation*}
which is a contradiction.

\subsection{Proof of Proposition \ref{prop: backward induction construction of Bellman optimal state-action policies}}
\label{app: proof 3}

\begin{itemize}
    \item We first prove that state-action policies $\Pi$ defined as in \eqref{eq: construction of Bellman optimal state-action policies} are Bellman optimal by induction on $T$.
    
    $T=1:$ 
    \begin{equation*}
         \Pi(a\mid s,0) >0 \iff a \in \arg\max_a \E[R(s_{1})\mid s_{0}=s,a_{0}=a], \quad \forall s\in \mathbb S
    \end{equation*}
    is a Bellman optimal state-action policy as it maximises the total reward possible in the MDP.
    
    Let $T>1$ be finite and suppose that the Proposition holds for MDPs with a temporal horizon of $T-1$. This means that
    
    \begin{equation*}
\begin{split}
 \Pi(a\mid s,T-1) >0 \iff a &\in \arg\max_a \E[R(s_{T})\mid s_{T-1}=s,a_{T-1}=a], \quad \forall s\in \mathbb S \\
   \Pi(a\mid s,T-2) >0 \iff a &\in \arg\max_a \E_{\Pi}[R(s_{T-1:T})\mid s_{T-2}=s,a_{T-2}=a], \quad \forall s\in \mathbb S \\
    &\vdots\\
  \Pi(a\mid s,1) >0 \iff a &\in \arg\max_a \E_{\Pi}[R(s_{2:T})\mid s_{1}=s,a_{1}=a] , \quad \forall s\in \mathbb S\\
  \end{split}
\end{equation*}
is a Bellman optimal state-action policy on the MDP restricted to times $1$ to $T$. Therefore, since
\begin{equation*}
\begin{split}
  \Pi(a\mid s,0) >0 \iff a &\in \arg\max_a \E_{\Pi}[R(s_{1:T})\mid s_{0}=s,a_{0}=a] , \quad \forall s\in \mathbb S\\
  \end{split}
\end{equation*}
Proposition \ref{prop: charac Bellman optimal state-action policy} allows us to deduce that $\Pi$ is Bellman optimal.

\item We now show that any Bellman optimal state-action policy satisfies the backward induction algorithm \eqref{eq: construction of Bellman optimal state-action policies}.

Suppose by contradiction that there exists a state-action policy $\Pi$ that is Bellman optimal but does not satisfy \eqref{eq: construction of Bellman optimal state-action policies}. Say, $\exists (a,s,t)\in \A \times \S\times \T, t<T$, such that
\begin{equation*}
\begin{split}
    \Pi(a\mid s,t) >0 \text{ and } a &\notin \arg\max_{\alpha\in \A} \E_\Pi[R(s_{t+1:T})\mid s_{t}=s,a_{t}=\alpha]. 
\end{split}
\end{equation*}
This implies
\begin{equation*}
    \E_\Pi[R(s_{t+1:T})\mid s_{t}=s,a_{t}=a] < \max_{\alpha\in \A} \E_\Pi[R(s_{t+1:T})\mid s_{t}=s,a_{t}=\alpha].
\end{equation*}

Let $\tilde a \in \arg\max_\alpha \E_\Pi[R(s_{t+1:T})\mid s_{t}=s,a_{t}=\alpha]$.
Let $\tilde \Pi$ be a state-action policy such that $\tilde \Pi(\cdot \mid s,t)$ assigns $\tilde a \in \A$ deterministically, and such that $\tilde \Pi = \Pi$ otherwise. Then we can contradict the Bellman optimality of $\Pi$ as follows

\begin{equation*}
\begin{split}
    v_{\Pi}(s,t) &= E_{\Pi}[R(s_{t+1:T})\mid s_t=s] \\
    &= \sum_{\alpha \in \A } E_{\Pi}[R(s_{t+1:T})\mid s_t=s, a_t=\alpha] \Pi(\alpha\mid s,t) \\
    &< \max_{\alpha\in \A} \E_\Pi[R(s_{t+1:T})\mid s_{t}=s,a_{t}=\alpha] \\
    &=\E_\Pi[R(s_{t+1:T})\mid s_{t}=s,a_{t}=\tilde a]\\
    &=\E_{\tilde \Pi}[R(s_{t+1:T})\mid s_{t}=s,a_{t}=\tilde a]\\
    &= \sum_{\alpha \in \A } E_{\tilde \Pi}[R(s_{t+1:T})\mid s_t=s, a_t=\alpha] \tilde \Pi (\alpha\mid s,t) \\
    &= v_{\tilde \Pi}(s,t).
\end{split}
\end{equation*}
\end{itemize}

\subsection{Proof of Lemma \ref{lemma: reward maxim EFE minim zero temp}}
\label{app: proof 4}

\begin{equation*}
\begin{split}
    &\lim_{\beta \to +\infty} \arg\min_{\vec a} \operatorname{D_{KL}}[Q(\vec s\mid \vec a,s_t) \mid C_\beta (\vec s)] \\
    &=\lim_{\beta \to +\infty} \arg \min_{\vec a} - \H[Q(\vec s\mid \vec a,s_t)]+\E_{Q(\vec s\mid \vec a,s_t)}[-\log C_\beta (\vec s)] \\
     &=\lim_{\beta \to +\infty} \arg \min_{\vec a} - \H[Q(\vec s\mid \vec a,s_t)]- \beta \E_{Q(\vec s\mid \vec a,s_t)}[ R(\vec s)] \\
    &=\lim_{\beta \to +\infty} \arg \max_{\vec a}  \H[Q(\vec s\mid \vec a,s_t)]+ \beta \E_{Q(\vec s\mid \vec a,s_t)}[ R(\vec s)] \\
    &\subseteq \lim_{\beta \to +\infty} \arg \max_{\vec a} \beta \E_{Q(\vec s\mid \vec a,s_t)}[ R(\vec s)] \\
    &= \arg \max_{\vec a} \E_{Q(\vec s\mid \vec a , s_t)}[R(\vec s)]
\end{split}
\end{equation*}

The inclusion follows from the fact that, as $\beta \to +\infty$, a minimiser of the expected free energy has to maximise $\E_{Q(\vec s\mid \vec a,s_t)}[ R(\vec s)]$. Among such action sequences, the expected free energy minimisers are those that maximise the entropy of future states $\H[Q(\vec s\mid \vec a,s_t)]$.

\subsection{Proof of Theorem \ref{thm: Bellman optimal t1}}
\label{app: proof 5}

 When $T=1$ the only action is $a_0$. We fix an arbitrary initial state $s_0 = s\in \S$. By Proposition \ref{prop: charac Bellman optimal state-action policy}, a Bellman optimal state-action policy is fully characterised by an action $a^*_0$ that maximises immediate reward
 \begin{equation*}
    a^*_0 \in \arg\max_{a\in \A} \E[R(s_{1})\mid s_0=s,a_0=a].
\end{equation*}
Recall that by Remark \ref{rem: notation expectation}, this expectation stands for return under the transition probabilities of the MDP
 \begin{equation*}
    a^*_0 \in \arg\max_{a\in \A} \E_{P(s_1\mid a_0=a, s_0=s)}[R(s_{1})].
\end{equation*}

Since transition probabilities are assumed to be known \eqref{eq: transition probabilities are known}, this reads
 \begin{equation*}
    a^*_0 \in \arg\max_{a\in \A} \E_{Q(s_1\mid a_0=a, s_0=s)}[R(s_{1})].
\end{equation*}

On the other hand, 

\begin{equation*}
\begin{split}
    a_0 &\in \lim_{\beta \to +\infty} \arg\max_{a \in \A} \exp (-G(a \mid s_t)) \\
    &= \lim_{\beta \to +\infty} \arg\min_{a \in \A} G(a \mid s_t).  
\end{split}
\end{equation*}

By Lemma \ref{lemma: reward maxim EFE minim zero temp}, this implies 
\begin{equation*}
    a_0 \in \arg\max_{a\in \A}\E_{Q(s_1\mid a_0=a, s_0=s)}[R(s_1)],
\end{equation*}
which concludes the proof.

\subsection{Proof of Theorem \ref{thm: backward induction as active inference}}
\label{app: proof 6}

We prove this result by induction on the temporal horizon $T$ of the MDP.

The proof of the Theorem when $T=1$ can be seen from the proof of Theorem \ref{thm: Bellman optimal t1}. Now suppose that $T>1$ is finite and that the Theorem holds for MDPs with a temporal horizon of $T-1$.

Our induction hypothesis says that $Q(a_\tau\mid s_\tau)$, as defined in \eqref{eq: sophisticated AI scheme theorem}, is a Bellman optimal state-action policy on the MDP restricted to times $\tau=1, \ldots, T$. Therefore, by Proposition \ref{prop: charac Bellman optimal state-action policy}, we only need to show that the action $a_0$ selected under active inference satisfies
\begin{equation*}
    a_0 \in \arg\max_{a\in \A} \E_{Q}[R(\vec s)\mid s_0, a_0=a].
\end{equation*}
This is simple to show as 
\begin{equation*}
    \begin{split}
        &\arg\max_{a\in \A} \E_{Q}[R(\vec s)\mid s_0, a_0=a] \\
        &= \arg\max_{a\in \A} \E_{P(\vec s\mid a_{1:T},a_0=a,s_0)Q(\vec a\mid s_{1:T})}[R(\vec s)] \quad \text{(by Remark \ref{rem: notation expectation})}\\
        &= \arg\max_{a\in \A} \E_{Q(\vec s,\vec a\mid a_0=a,s_0)}[R(\vec s)] \quad \text{(as the transitions are known)}\\
        &= \lim_{\beta \to +\infty} \arg\max_{a\in \A} \E_{Q(\vec s,\vec a\mid a_0=a,s_0)}[\beta R(\vec s)]\\
        &\supseteq \lim_{\beta \to +\infty} \arg\max_{a\in \A} \E_{Q(\vec s,\vec a\mid a_0=a,s_0)}[\beta R(\vec s)] -\H[Q(\vec s\mid \vec a,a_0=a,s_0)]\\
        &= \lim_{\beta \to +\infty} \arg\min_{a\in \A} \E_{Q(\vec s,\vec a\mid a_0=a,s_0)}[-\log C_\beta(\vec s)] - \H[Q(\vec s\mid \vec a,a_0=a,s_0)] \quad \text{ (by \eqref{eq: defining preferences as reward maximising})}\\
        &=\lim_{\beta \to +\infty} \arg\min_{a\in \A} \E_{Q(\vec s,\vec a\mid a_0=a,s_0)}\operatorname{D_{KL}}[Q(\vec s\mid \vec a, a_0=a,s_0)\mid  C_\beta(\vec s)]\\
        &=\lim_{\beta \to +\infty} \arg \min_{a\in \A} G(a_0=a\mid s_0) \quad \text{(by \eqref{eq: unpacking recursive EFE})}.
    \end{split}
\end{equation*}
Therefore, an action $a_0$ selected under active inference is a Bellman optimal state-action policy on finite temporal horizons. Furthermore, the inclusion follows from the fact that if there are multiple actions that maximise expected reward, that which is selected under active inference maximises the entropy of beliefs about future states.

\subsection{Proof of Proposition \ref{prop: reward maxim EFE minim zero temp POMDPs}}
\label{app: proof 7}

Unpacking the zero temperature limit
\begin{equation*}
\begin{split}
    &\lim_{\beta \to +\infty} \arg\min_{\vec a } G(\vec a\mid \tilde o) \\
    &=\lim_{\beta \to +\infty} \arg\min_{\vec a} \operatorname{D_{KL}}[Q(\vec s\mid \vec a,\tilde o)\mid C_\beta(\vec s)]+\E_{Q(\vec s\mid \vec a,\tilde o)}\H [P(\vec o\mid \vec s)]\\
    &=\lim_{\beta \to +\infty} \arg \min_{\vec a} - \H[Q(\vec s\mid \vec a,\tilde o)]+\E_{Q(\vec s\mid \vec a,\tilde o)}[-\log C_\beta (\vec s)]+\E_{Q(\vec s\mid \vec a,\tilde o)}\H [P(\vec o\mid \vec s)] \\
    &=\lim_{\beta \to +\infty} \arg \min_{\vec a} - \H[Q(\vec s\mid \vec a,\tilde o)]- \beta \E_{Q(\vec s\mid \vec a,\tilde o)}[ R(\vec s)]+\E_{Q(\vec s\mid \vec a,\tilde o)}\H [P(\vec o\mid \vec s)] \quad \text{(by \eqref{eq: defining preferences as reward maximising})} \\
    &\subseteq \lim_{\beta \to +\infty} \arg \max_{\vec a} \beta \E_{Q(\vec s\mid \vec a,\tilde o)}[ R(\vec s)] \\
    &= \arg \max_{\vec a} \E_{Q(\vec s\mid \vec a , \tilde o)}[R(\vec s)]
\end{split}
\end{equation*}

The inclusion follows from the fact that as $\beta \to +\infty$ a minimiser of the expected free energy has first and foremost to maximise $\E_{Q(\vec s\mid \vec a,\tilde o)}[ R(\vec s)]$. Among such action sequences, the expected free energy minimisers are those that maximise the entropy of (beliefs about) future states $\H[Q(\vec s\mid \vec a,\tilde o)]$ and resolve ambiguity about future outcomes by minimising $\E_{Q(\vec s\mid \vec a,\tilde o)}\H [P(\vec o\mid \vec s)]$.

\end{document}